\documentclass[letterpaper]{article} 
\usepackage{aaai25}  
\usepackage{times}  
\usepackage{helvet}  
\usepackage{courier}  
\usepackage[hyphens]{url}  
\usepackage{graphicx} 
\usepackage{natbib}  
\usepackage{caption} 
\usepackage{booktabs}
\usepackage{algorithm}
\usepackage{algorithmic}
\usepackage{amsmath}
\usepackage{amssymb}
\usepackage{xcolor}
\usepackage{colortbl}
\usepackage{multirow}
\usepackage{graphicx, subcaption, overpic, textpos}
\usepackage{newfloat}
\usepackage{listings}
\DeclareCaptionStyle{ruled}{labelfont=normalfont,labelsep=colon,strut=off} 
\floatstyle{ruled}
\newfloat{listing}{tb}{lst}{}
\floatname{listing}{Listing}

\definecolor{bblue}{rgb}{0,150,230}
\definecolor{mygray}{gray}{.9}
\definecolor{lightgray}{gray}{.96}
\definecolor{myy}{RGB}{126,95,0}
\definecolor{ggray}{RGB}{127,127,127}
\definecolor{mygreen}{RGB}{93,173,85}
\definecolor{myred}{RGB}{240,16,89}
\definecolor{myblue}{RGB}{0,114,188}
\definecolor{darkgreen}{rgb}{0.0, 0.5, 0.0}
\definecolor{demphcolor}{RGB}{100,100,100}

\makeatletter
\newcommand{\thickhline}{%
    \noalign {\ifnum 0=`}\fi \hrule height 0.8pt
    \futurelet \reserved@a \@xhline
}
\makeatother

\newlength\savewidth\newcommand\shline{\noalign{\global\savewidth\arrayrulewidth
  \global\arrayrulewidth 1pt}\hline\noalign{\global\arrayrulewidth\savewidth}}
\newcommand{\tablestyle}[2]{\setlength{\tabcolsep}{#1}\renewcommand{\arraystretch}{#2}\centering\footnotesize}
\renewcommand{\paragraph}[1]{\vspace{1.25mm}\noindent\textbf{#1}}

\newcommand{\app}{\raise.17ex\hbox{$\scriptstyle\sim$}}

\definecolor{deemph}{gray}{0.6}

\definecolor{baselinecolor}{gray}{.9}
\newcommand{\baseline}[1]{\cellcolor{baselinecolor}{#1}}

\title{Restorer: Removing Multi-Degradation with All-Axis Attention and Prompt Guidance}
\author{
    Jiawei Mao\textsuperscript{1}, Juncheng Wu\textsuperscript{2}, Yuyin Zhou\textsuperscript{2}, Xuesong Yin\textsuperscript{1}, Yuanqi Chang\textsuperscript{1}
}
\affiliations{
    \textsuperscript{1}Hangzhou Dianzi University,      \textsuperscript{2}UC Santa Cruz\\

}

\usepackage{bibentry}

\begin{document}

\maketitle


\begin{abstract}
There are many excellent solutions in image restoration.
However, most methods require on training separate models to restore images with different types of degradation.
Although existing all-in-one models effectively address multiple types of degradation simultaneously, their performance in real-world scenarios is still constrained by the task confusion problem.
In this work, we attempt to address this issue by introducing \textbf{Restorer}, a novel Transformer-based all-in-one image restoration model.
To effectively address the complex degradation present in real-world images, we propose All-Axis Attention (AAA), a novel attention mechanism that simultaneously models long-range dependencies across both spatial and channel dimensions, capturing potential correlations along all axes.
Additionally, we introduce textual prompts in Restorer to incorporate explicit task priors, enabling the removal of specific degradation types based on user instructions. By iterating over these prompts, Restorer can handle composite degradation in real-world scenarios without requiring additional training.
Based on these designs, Restorer with one set of parameters demonstrates state-of-the-art performance in multiple image restoration tasks compared to existing all-in-one and even single-task models.
Additionally, Restorer is efficient during inference, suggesting the potential in real-world applications. Code will be available at \textbf{https://github.com/Talented-Q/Restorer}.
\end{abstract}

%

 \begin{figure}[t]
    \setlength{\belowcaptionskip}{-7mm}
    \centering
    \includegraphics[width=1\linewidth]{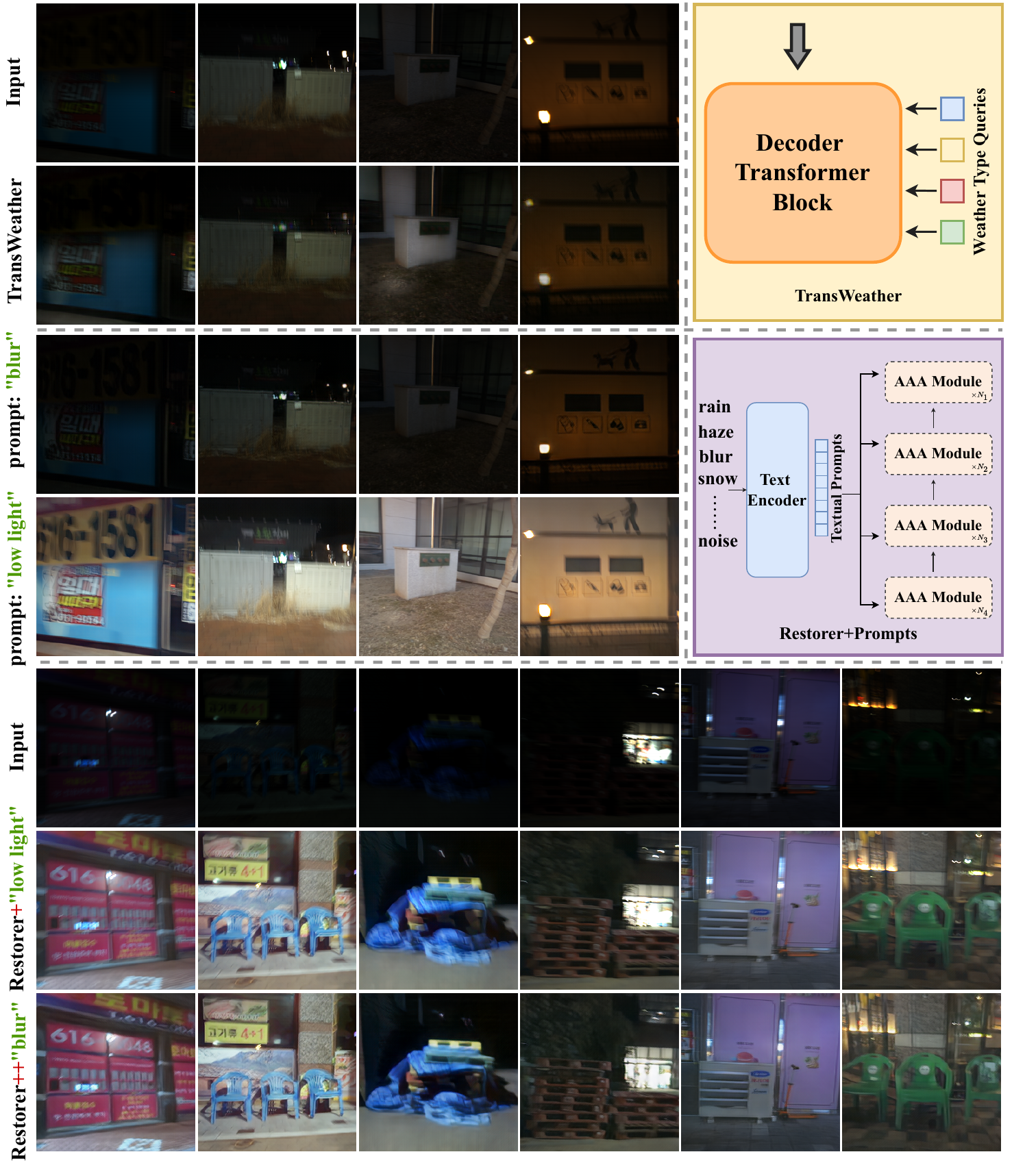}
     \caption{Restored images of TransWeather and Restorer based on different textual prompts. TransWeather confuses the low-light enhancement task when performing the deblurring task, resulting in poor deblurring results. In contrast, Restorer at different textual prompts accurately performs the corresponding image restoration task.}
 \label{fig131}
  \end{figure}

\section{Introduction}



 \begin{figure*}[!t]
    \setlength{\abovecaptionskip}{0.5mm} 
    \setlength{\belowcaptionskip}{-6mm}
    \centering
    \includegraphics[width=0.75\linewidth]{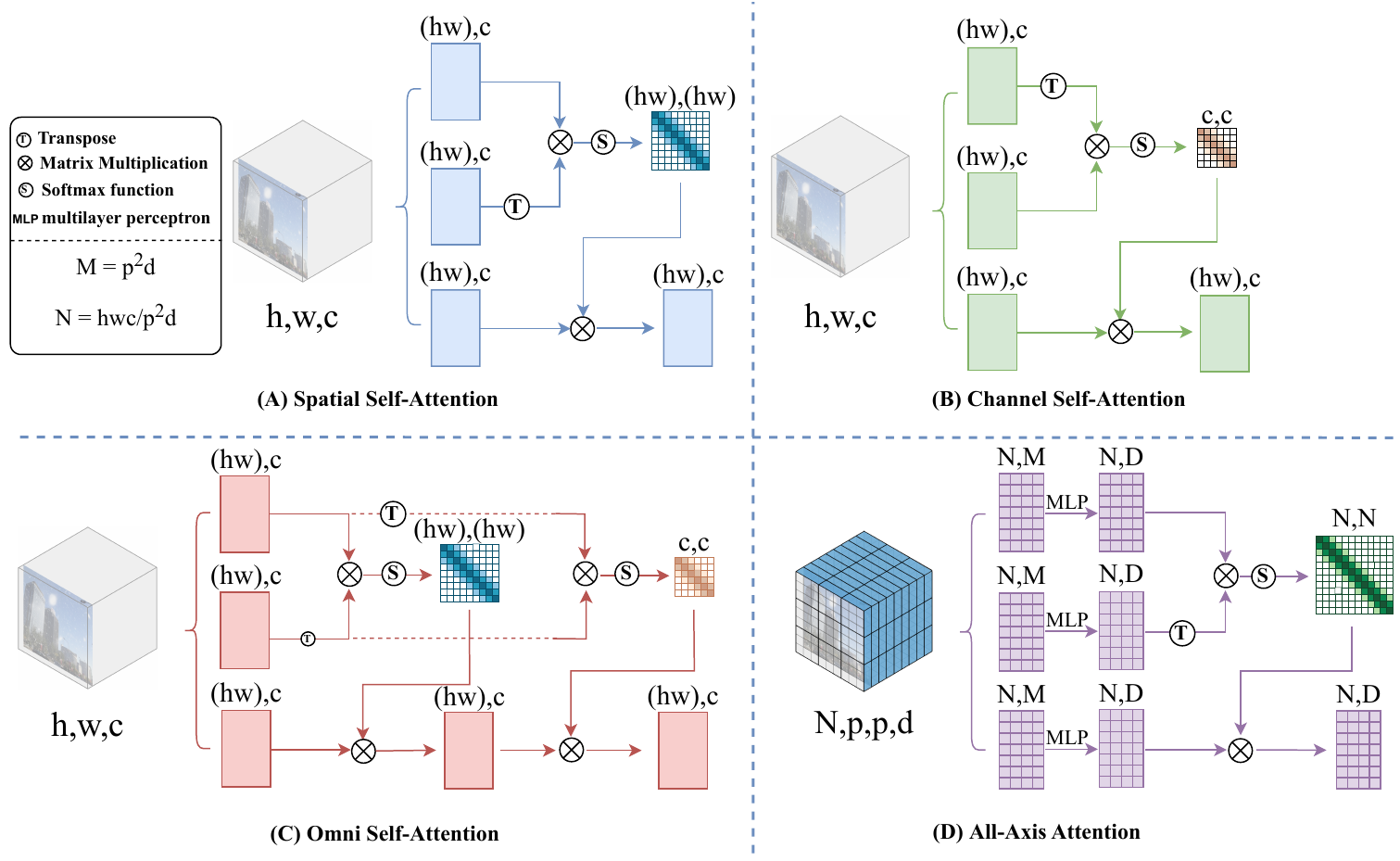}
     \caption{Illustration of spatial self-attention, channel self-attention, omni self-attention, and all-axis attention.}
 \label{fig2}
  \end{figure*}

 Environments such as severe weather (rain, fog, and snow) and low light can reduce image visibility. Cameras produce noise and blurring issues when shooting. 
 All these negative factors dramatically affect many computer vision works.
 To tackle this challenge, various image restoration algorithms including image deraining~\cite{ren2019progressive,ye2021closing,fu2021rain}, desnowing~\cite{liu2018desnownet,hdcwnet,zhang2021deep}, defogging~\cite{wu2021contrastive,qin2020ffa,dong2020multi}, deblurring~\cite{park2020multi,zhang2020deblurring,kim2022mssnet}, denoising~\cite{ren2021adaptive,zhang2017beyond,guo2019toward}, and low-light image enhancement~\cite{fan2022half,wang2018gladnet} have been widely explored nowadays. 
 Despite their promising performance, they lack the generalization ability to different types of degradation, restricting their application in real-world scenario, where images may contain multiple types of unknown degradation.
 
 A straight-forward idea to address this issue is to switch between a range of image restoration models~\cite{li2020all} according to the degradation type. However, this is undoubtedly costly.
\citet{li2022all} proposed an all-in-one image restoration model AirNet, enabling removal of multiple types of degradation with one set of parameters by learning the degradation representation through contrastive learning.
 While \citet{chen2022learning} proposed to discriminate and remove multiple adverse weather through two-stage knowledge distillation and multi-contrastive regularization.
 However, recent research~\cite{potlapalli2023promptir} points out that contrastive learning-based methods may fail to extract fully disentangled representations for different degradation types. 
 An alternative method to model degradation representations is to employ learnable degradation embedding~\cite{valanarasu2022transweather,potlapalli2023promptir}.
 These embeddings are learned in an end-to-end manner and interacted with image features to provide degradation information.
 Although these methods achieve encouraging results in different types of restoration tasks, their performance on real-world images containing composite degradation is still limited by the task confusion problem (see Figure~\ref{fig131}).

 In this work, we aim to address above issues in all-in-one image restoration task. 
 Specifically, to effectively extract image features containing rich degradation information, we propose a novel All-Axis Attention (AAA) module to concurrently model long-range dependencies across both spatial and channel dimensions.
 Moreover, we introduce textual degradation prompt to guide degradation removal, further addressing the issue of task confusion.

 Attention modeling is crucial for Transformer-based image restoration models to extract high-quality features.
 Mainstream image restoration models~\cite{valanarasu2022transweather,tsai2022stripformer,song2023vision} mainly use spatial attention scheme.
 However, recent researches~\cite{zamir2022restormer,wang2023omni} show that attention modeling in channel dimensions is also crucial for image restoration.
 Based on this observation,~\citet{wang2023omni} proposed Omni Self-Attention (OSA), consisting of two-stage attention modeling of spatial and channel, achieving impressive performance on image super-resolution.
 However, the two-stage asynchronous attention modeling is unable to fully exploit the potential dependencies between all axis in the degraded image, restricting the ability of OSA in effectively modeling of degradation representations.
 Therefore, in this paper, we propose the AAA module with stereo embedding and 3D convolution.
 Our AAA extends the feature interaction to 3D space, synchronizing the attention modeling in spatial and channel dimensions, making it more suitable for handling multiple image restoration tasks.

  \begin{figure*}[!t]
    \centering
    \setlength{\belowcaptionskip}{-5mm}
    \includegraphics[width=0.93\linewidth]{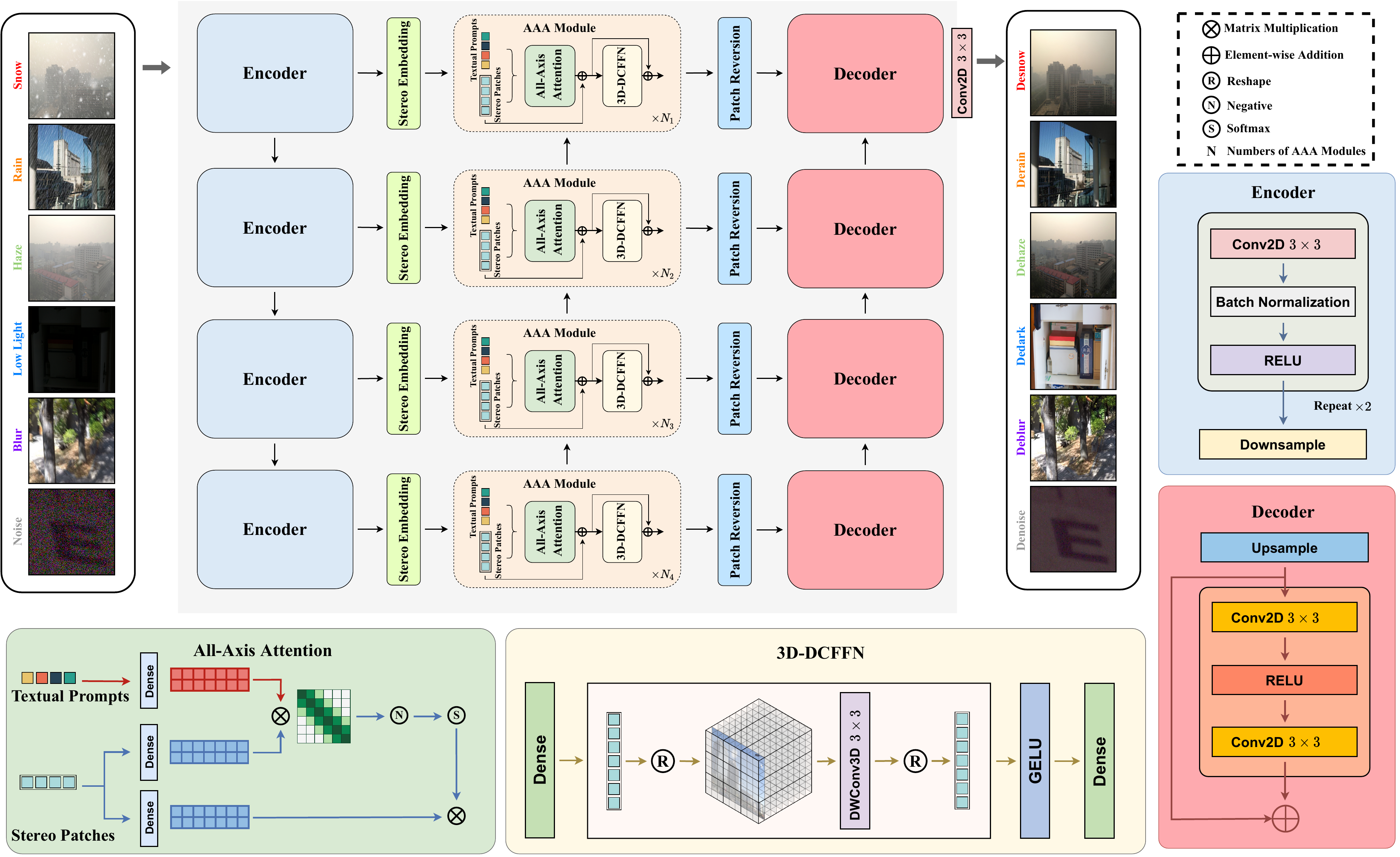}
     \caption{The overall architecture of the proposed Restorer and the structure of each Restorer component. We utilize transposed convolution for upsampling and convolution with a stride of 2 for downsampling, respectively.}
     \label{fig3}
   \end{figure*}

 Additionally, to precisely remove degradation from image features.
 We introduce textual degradation prompts as queries for All-Axis Attention modeling, explicitly providing degradation information according to user instructions.
 Compared to learnable prompt embeddings that lack interpretability, textual prompts offer deterministic task prioritization, enabling precise removal of the corresponding degradation based on guidance from the textual prompt embeddings.
 Meanwhile, for images with multiple degradations, we can simply remove composite degradations by iterating over different degradation prompts.

 Integrating All-Axis Attention module and textual degradation prompts, we propose the \textbf{Restorer}, a Transformer-based all-in-one image restoration model with the U-Net~\cite{ronneberger2015u} architecture, efficiently addressing multiple image restoration tasks. 
 Experiments on numerous standard datasets show that Restorer achieves either state-of-the-art or competitive results on multiple image restoration tasks compared to not only universal image restoration models, but also models specifically designed for each individual tasks. 
 Moreover, the results on the real-world dataset demonstrate the robustness of Restorer against various types of degradation, demonstrating the potential of Restorer as a practical solution for image restoration tasks.

 \section{Related Work}



 \subsection{All-in-One Image Restoration}

 ~\citet{li2020all} proposed All in One for image restoration with a scheme of neural architecture search and adversarial training. However, All in One requires multiple specific encoders for each types of degradation, restricting its efficiency. 
 To address this problem, TKL~\cite{chen2022learning} learns crucial features for different types of degradation from many encoders through two-stage knowledge distillation and multi-contrast knowledge regularization.
 While AirNet~\cite{li2022all} removes various types of degradation using a unified model by contrastive-based degradation encoder and degradation-guided restoration model.
 Furthermore, PromptIR~\cite{potlapalli2023promptir} employs learnable parameters as prompts to encode key discriminative information about various types of degradation. 
 And MPerceiver~\cite{ai2024multimodal} exploits the stable diffusion prior to enhancing the adaptability, versatility, and fidelity of all-in-one image restoration.
 In this paper, we propose AAA module to synchronously modeling degradation representation in both spatial and channel dimensions.
 Our experimental results show that the proposed AAA module is better suited to the requirements of all-in-one image restoration task.
 
  \subsection{Prompt Learning}
Prompt learning is first proposed in Natural Language Processing filed. Pre-trained models perform the target task based on manual prompts~\cite{petroni2019language,brown2020language}. 
Many of the follow-up efforts tried to employ discrete~\cite{shin2020autoprompt,wallace2019universal} or continuous~\cite{li2021prefix,lester2021power} automated prompts to reduce labor costs of designing prompts. 
Meanwhile, prompt learning has been introduced to computer vision and has become a key component in various visual models~\cite{zhou2022learning,ge2023domain}. 
VPT~\cite{jia2022visual} designed visual prompts. Notably, SAM~\cite{kirillov2023segment} unified the semantic segmentation task by leveraging multiple prompt types. 
Depending on the flexibility of the prompt, MAE-VQGAN~\cite{bar2022visual} and Painter~\cite{wang2023images} unified several visual tasks. Transweather~\cite{valanarasu2022transweather} proposes learnable weather type embedding as prompts to adapt multiple severe weather degradation.
However, the lack of interpretability of learnable embeddings as prompts may lead to all-in-one image restoration methods suffering from task confusion, especially in composite degradation scenes. 
In contrast, Restorer with All-Axis Attention modeling and interpretable textual prompts, presents a promising solution to this issue.

  \begin{figure*}[!t]
    \centering
    \setlength{\belowcaptionskip}{-5mm}
    \includegraphics[width=0.85\linewidth]{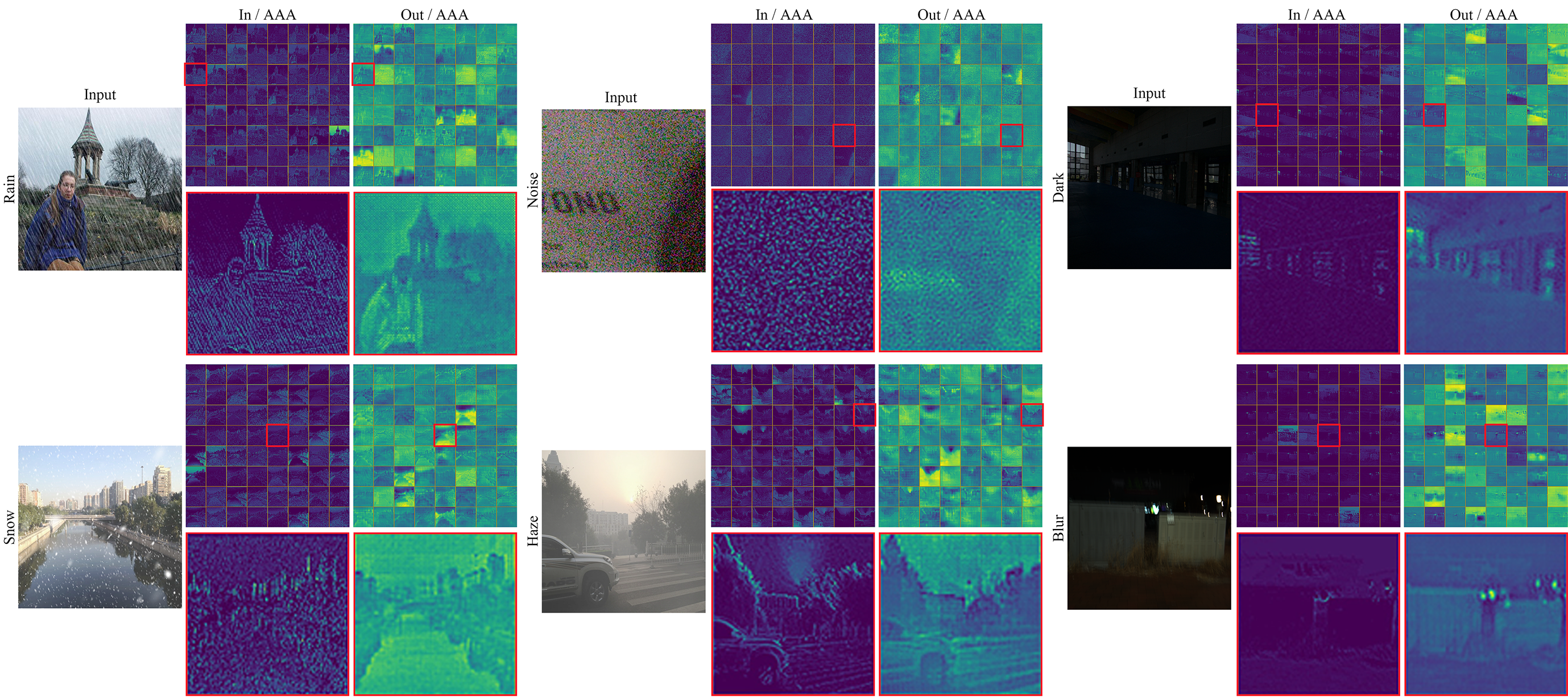}
     \caption{Visualization of feature maps for each channel before and after the AAA module for different image restoration tasks. It can be observed that there are significant degradation residuals in the inter-channel feature maps before the AAA module, while the degradation residuals among the channels are eliminated after the AAA module.}
     \label{fig4}
   \end{figure*}

 \section{Proposed Method}

 Our main goal is to develop an efficient all-in-one model that can be applied to multiple image restoration tasks. In this section, we first describe the overall pipeline of Restorer (see Figure~\ref{fig3}). Then, we discuss the core components in Restorer, including the stereo embedding, all-axis attention modeling, textual prompts, and 3D deep convolutional feed-forward networks. 

 \subsection{Overall Pipeline}

 Firstly, we aim to extract features from the degraded image using four-levels of encoder.
 As shown in Figure~\ref{fig3}, Restorer's encoder uses a compact convolutional architecture to extract low-level feature embeddings for computational efficiency.
 For different stages of the encoder output, we transform it into a sequence of stereo token embeddings.
 Then, All-Axis-Attention (AAA) module models potential dependencies between textual prompts and the degraded feature in the all-axis, and filters out the degenerate features from the feature embedding in both spatial and channel dimensions via negative affinity matrices. 
 Next, the AAA module performs fine-grained feature extraction of stereo patches through 3D-DCFFN to preserve the high-frequency information in the image.
 For the stereo patch output from the AAA module, the convolutional decoder converts them into feature maps through the patch reversion layer and progressively restores a high-resolution representation by stacked convolution and transposed convolution while preserving the fine details of the image.
 The design of the encoder-AAA-decoder in different stages guarantees Restorer's ability to effectively capture visual information at different scales. 
 Eventually, a high-fidelity restored image is obtained from the $3\times3$ convolutional projection block.

 \subsection{All-Axis Attention Module}
 
\begin{figure*}[!t]
    \centering
    \setlength{\abovecaptionskip}{1mm} 
    \setlength{\belowcaptionskip}{-2mm}
    \includegraphics[width=1\linewidth]{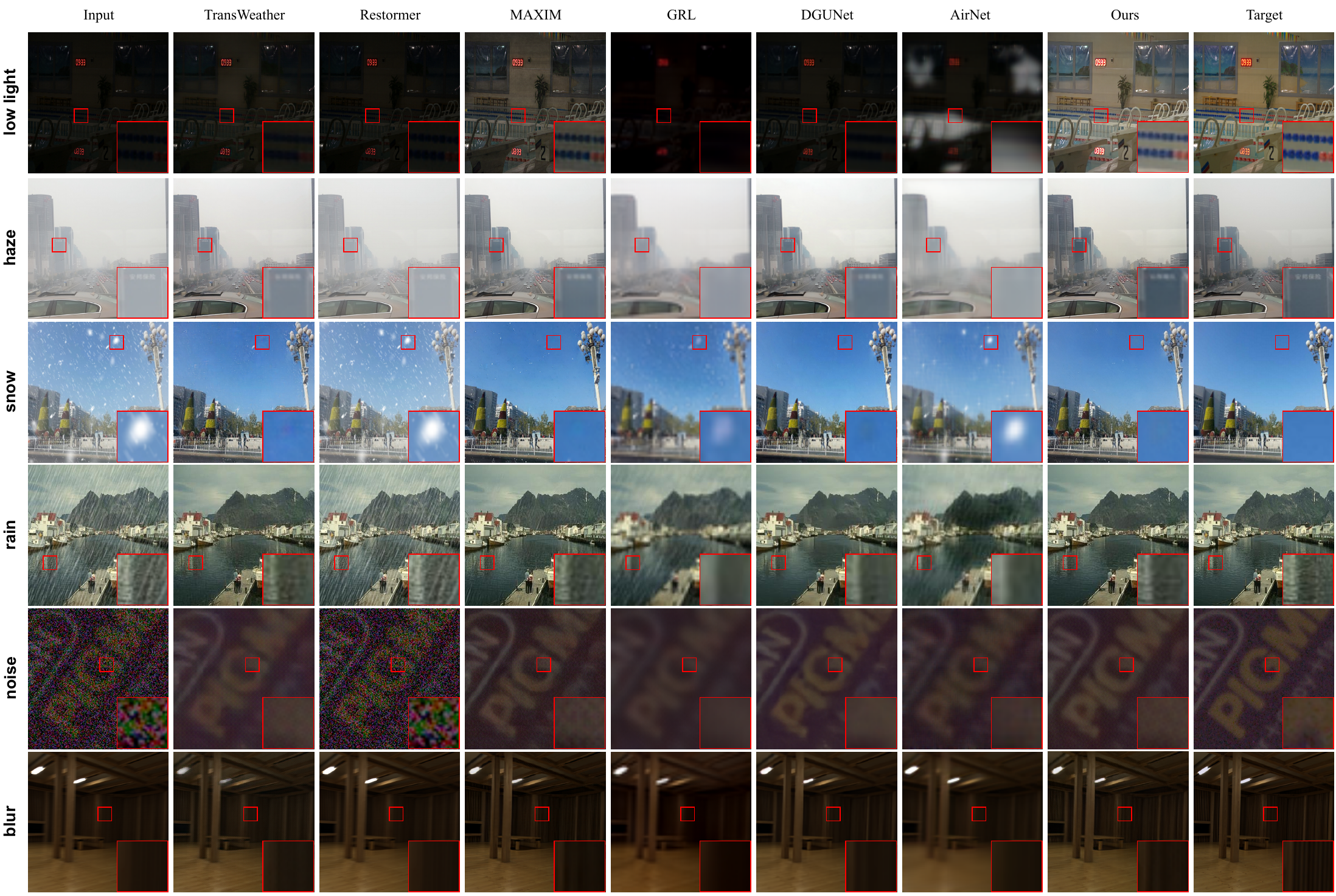}
     \caption{Visual comparison with SOTA image restoration algorithms in different image restoration tasks.}
     \label{fig5}
   \end{figure*}

 As illustrated in Figure~\ref{fig2}, the one-dimensional attention modeling is limited in spatial or channel dimension, thus fails to exploit the full potential of the attention mechanism in image restoration tasks. 
 For the spatial attention operator~\cite{chen2021pre,vaswani2017attention}, since the channels share the same spatial weights, it can only mine potential correlations in the degraded image space, ignoring the explicit use of channel information. 
 Recent studies~\cite{zamir2022restormer,wang2023omni} have shown that the channel attention operator is more compact than the spatial attention operator, and it also plays an important role in image restoration tasks. However, the channel attention operator ignores critical spatial information in the image restoration task leading to a drastic reduction in the relational modeling capability compromising the accuracy of aggregation (see Figure~\ref{fig5}), especially for unified image restoration tasks.
 Although~\citet{wang2023omni} proposed Omni self-attention (OSA) to address the limitation of the one-dimensional self-attention, this approach requires two stages of modeling through both spatial attention and channel attention. 
 This asynchrony prevents OSA from fully mining the potential dependencies between the all-axis of degraded images.
 In contrast, our All-Axis Attention extends the interaction to 3D space through stereo embedding, enjoying the advantages of spatial and channel attention modeling, making it more suitable for image restoration tasks.

\begin{table*}[!t]
\centering
\setlength{\abovecaptionskip}{0.1cm} 
\setlength{\belowcaptionskip}{-5mm}
\resizebox{17.8cm}{!}{
\setlength\tabcolsep{2pt}
\renewcommand\arraystretch{1.1}
\begin{tabular}{c||cc|cc|cc|cc|cc|cc|cc|cc|c}
\hline\thickhline
\rowcolor{mygray}
 &\multicolumn{2}{c|}{ \textbf{deraining} } &\multicolumn{2}{c|}{ \textbf{desnowing} } & \multicolumn{2}{c|}{ \textbf{defogging} } & \multicolumn{2}{c|}{\textbf{denoising} }& \multicolumn{2}{c|}{ \textbf{low-light enhancement} }& \multicolumn{4}{c|}{\textbf{deblurring} } & \multicolumn{2}{c|}{} & \\\cline{2-15}
\rowcolor{mygray}
 &\multicolumn{2}{c|}{ \textbf{rain1400} } &\multicolumn{2}{c|}{ \textbf{CSD} } & \multicolumn{2}{c|}{ \textbf{SOTS} } & \multicolumn{2}{c|}{\textbf{SIDD} }& \multicolumn{2}{c|}{ \textbf{LOL} }& \multicolumn{2}{c|}{\textbf{GoPro} }& \multicolumn{2}{c|}{\textbf{RealBlur-R} } & \multicolumn{2}{c|}{\multirow{-2}*{\textbf{Average}}} & \\\cline{2-17}
 \rowcolor{mygray}
  \multirow{-3}*{Method} & PSNR $\uparrow$ & SSIM($\%$) $\uparrow$ & PSNR $\uparrow$ & SSIM($\%$) $\uparrow$ & PSNR $\uparrow$ & SSIM($\%$) $\uparrow$ & PSNR $\uparrow$ & SSIM($\%$) $\uparrow$ & PSNR $\uparrow$ & SSIM($\%$) $\uparrow$ & PSNR $\uparrow$ & SSIM($\%$) $\uparrow$ & PSNR $\uparrow$ & SSIM($\%$) $\uparrow$ & PSNR $\uparrow$ & SSIM($\%$) $\uparrow$ & \multirow{-3}*{Time}  \\\cline{2-18}
GRL & 20.56 & 79.71 & 20.45 &80.01 &21.71& 84.97 & 33.81 & 93.53 & 7.09 & 3.82 & 18.50 & 77.94 & 27.58 & 49.15 & 21.42 & 67.01& 515ms \\
Uformer & 31.74 &95.49 & 27.46 &92.61 &29.85& 97.82 & 38.35 & 96.26 &8.50& 26.84&25.89&91.38&37.13&95.86&28.41&85.18& 35ms \\
DGUNet & 33.05 & 96.52 &28.81 & 92.28 & 31.04 & 96.22 & \textbf{39.61} & \textbf{97.04}&13.04& 55.37& \underline{30.06} &94.04&37.74&96.36&30.47&89.69& 38ms \\
MPRNet & 33.47 & 96.59 & 28.03 &92.57 &30.96& 97.40& \underline{38.94} & \underline{96.75} & 15.78&71.15 &28.39 & \underline{94.10} &37.50& 96.75 &30.43&92.18&32ms\\
Restormer & 24.38 & 87.60 &14.30 &72.67 &15.94 & 82.92 &25.30  & 65.35&7.79& 19.79&26.22 &87.15&34.42&\underline{96.83}&21.19&73.18 & 75ms\\
MAXIM &  32.63 & \underline{96.80} & 27.81 &92.20 &32.20& \underline{98.14} & 27.70 & 95.95&13.48&78.33&25.78&93.39&32.52&91.87&27.44&92.38& 71ms\\
NAFNet & 32.87 & \textbf{97.32} &28.31 &\underline{92.65}&32.34&96.74&29.89 &96.18& \textbf{21.57} & \underline{89.37}&25.58&93.46&32.91&95.18&29.06& \underline{94.41} & 24ms \\
\hline
TransWeather & \textbf{34.27} &96.25&\underline{29.38} &92.22&\underline{32.96}&96.55& 36.14&95.69&15.72&62.90&29.03&93.78& \underline{39.00} &93.06& \underline{30.92} &90.06& 14ms\\
SYENet & 30.02 & 89.11 & 22.55 & 87.35 & 26.37 & 95.66 & 33.19 & 91.03 & 17.64 & 81.57 & 26.78 & 93.30 & 35.90 & 94.92&27.49&90.42& 6ms \\
AirNet & 24.63 & 83.67 & 18.40 &72.28 &18.14&76.86&35.10 & 87.30&11.24&32.10&24.18&85.88&26.94&85.57&22.66&74.80& 49ms \\
PromptIR & 30.62 & 94.98 & 27.42 & 91.70 & 29.59 & 96.79 & 36.25 & 89.86 & 15.42 & 73.01 & 26.75 & 86.26 & 36.58 & 94.91 & 28.95 & 89.64 & 81ms\\ 
OneRestore & 25.56 & 89.98 & 28.80 & 92.00 & 22.35 & 91.18 & 35.17 & 92.94 & 15.13 & 78.18 & 28.19 & 93.68 & 37.06 & 95.16 & 27.46 & 90.44 & 17ms \\
\hline
Restorer (Ours) & \underline{34.09} &96.67&\textbf{30.28} & \textbf{93.09} &\textbf{34.46}& \textbf{98.25} &37.75 & 96.36 & \underline{21.44}& \textbf{90.53}&\textbf{30.92}&\textbf{95.48}&\textbf{43.10}&\textbf{97.55}&\textbf{33.14}&\textbf{95.41}& 22ms \\
\hline\thickhline
\end{tabular}
}
\caption{Quantitative comparison with image restoration baselines. Time denotes the inference time.}
\label{tab1}
\end{table*}

 \paragraph{Stereo Embedding.} 
 For the AAA module in different stages, Restorer employs stereo embedding to divide the feature map $ \rm{F} \in {\mathbb{R}^{\rm{h}\times\rm{w}\times\rm{c}}}$ into stereo token embedding, where $\rm{h}\times\rm{w} $ denotes the spatial dimension and $\rm{c}$ denotes the number of channels. Specially, the stereo embedding first reshapes the feature map into a sequence of stereo tokens $ \rm{E} \in {\mathbb{R}^{\rm{p}\times\rm{p}\times\rm{d}\times(\rm{n_p^2n_d})}}$, where $ \rm{p^2} $ and $ \rm{d} $ denote the stereo token space dimension and the channel dimension respectively, and $ \rm{N}=\rm{n_p}\times\rm{n_p}\times\rm{n_d}=\rm{hwc}/\rm{p^2d} $ represents the resulting number of stereo tokens. 
 The stereo embedding then replaces the time dimension in the 3D convolution with the channel dimension to map the stereo token, thus introducing stereo spatial information for each stereo token.
 Next, we flatten the stereo token sequence and project it to a constant potential vector size D with linear layer to obtain the stereo token embedding $ \rm{E} \in {\mathbb{R}^{\rm{N}\times\rm{D}}}$ following~\cite{dosovitskiy2020image}.
 To stabilize the learning process, layer normalization~\cite{ba2016layer} is applied for stereo token embedding. Finally, we add learnable position embedding $\rm{E_{pos}}$ for stereo token embedding to preserve stereo position information (see Equation~\ref{eq10}).
 Compared to other token embedding division methods, stereo token embedding divided in 3D space allow for synchronized modeling of long-range dependencies in both space and channel through self-attention. 
 \begin{equation}
   \begin{split}
         &{\rm{F}} = [{\rm{f^{(1)}}};{\rm{f^{(2)}}};...;{\rm{f^{(i)}}};...;{\rm{f^{(N)}}}],{\rm{f^{(i)}}} \in {\mathbb{R}^{\rm{p}\times\rm{p}\times\rm{d}}},\\ 
         &{\rm{Z}} = \rm{Flat(3DConv(F))},\\
         &{\rm{Z}} = [{\rm{z^{(1)}W}};{\rm{z^{(2)}W}};...;{\rm{z^{(i)}W}};...;{\rm{z^{(N)}W}}],\\&{\rm{z^{(i)}}} \in {\mathbb{R}^{\rm{p^2d}}},{\rm{W}} \in {\mathbb{R}^{\rm{(p^2d)}\times\rm{D}}},\\
         &{\rm{Z}} = {\rm{LN(Z)+E_{pos}}},{\rm{E_{pos}}} \in {\mathbb{R}^{\rm{N}\times\rm{D}}},
  \end{split}
   \label{eq10}
 \end{equation}
where $\rm{3DConv(.)}$ denotes 3D convolution, $\rm{Flat(.)}$ is the flattening operation, $\rm{W}$ is the linear layer weights, and $\rm{LN(.)}$ is the layer normalization operation.

\begin{table*}[t]
\vspace{-.2em}
\footnotesize
\centering
\subfloat[
\textbf{Deraining}. rain1400 draining results.
\label{tab:deraining}
]{
\centering
\begin{minipage}{0.4\linewidth}{\begin{center}
\tablestyle{4pt}{1.05}
\tiny
\begin{tabular}{c|c|c}
Method & PSNR & SSIM ($\%$) \\
\shline
JORDER~\cite{yang2017deep} & 31.28 & 92.00 \\
PReNet~\cite{ren2019progressive} & 31.88 & 93.00 \\
DRD-Net~\cite{deng2020detail} & 29.65 & 88.00 \\
MSPFN~\cite{jiang2020multi} & 29.24 & 88.00 \\
EfficientDeRain~\cite{guo2021efficientderain} & 32.30 & 92.72 \\
JRGR~\cite{ye2021closing} & 31.18 & 91.00 \\
RCDNet~\cite{wang2020model} & \underline{33.04} & \underline{94.72} \\
\baseline{Restorer (Ours)} & \baseline{\textbf{34.09}} & \baseline{\textbf{96.67}}
\end{tabular}
\footnotesize
\end{center}}\end{minipage}
}
\hspace{2em}
\subfloat[
\textbf{Defogging}. SOTS defogging results.
\label{tab:defogging}
]
{
\begin{minipage}{0.4\linewidth}{\begin{center}
\tablestyle{4pt}{1.05}
\tiny
\begin{tabular}{c|c|c}
Method & PSNR & SSIM ($\%$) \\
\shline
EPDN~\cite{qu2019enhanced} & 23.82 & 87.00 \\
PFDN~\cite{dong2020physics} & 31.45 & 97.00 \\
KDDN~\cite{hong2020distilling} & 29.16 & 94.00 \\
MSBDN~\cite{dong2020multi} & 33.79 & 98.00 \\
FFA-Net~\cite{qin2020ffa} & \underline{34.98} & \textbf{99.00} \\
AECRNet~\cite{wu2021contrastive} & \textbf{35.61} & 98.00 \\
DehazeFormer~\cite{song2023vision} & 34.29 & \underline{98.30} \\
\baseline{Restorer (Ours)} & \baseline{34.46} & \baseline{98.25}
\end{tabular}
\end{center}}\end{minipage}
}
\hspace{2em}
\subfloat[
\textbf{Desnowing}. CSD desnowing results.
\label{tab:desnowing}
]{
\begin{minipage}{0.4\linewidth}{\begin{center}
\tablestyle{4pt}{1.05}
\tiny
\begin{tabular}{c|c|c}
Method & PSNR & SSIM ($\%$) \\
\shline
DesnowNet~\cite{liu2018desnownet} & 20.13 & 81.00 \\
CycleGAN~\cite{zhu2017unpaired} & 20.98 & 80.00 \\
JSTASR~\cite{chen2020jstasr} & 27.96 & 88.00 \\
DDMSNet~\cite{zhang2021deep} & 27.24 & 82.00 \\
HDCW-Net~\cite{hdcwnet} & \underline{29.06} & \underline{91.00} \\
DesnowGAN~\cite{jaw2020desnowgan} & 28.63 & 90.00 \\
\baseline{Restorer (Ours)} & \baseline{\textbf{30.28}} & \baseline{\textbf{93.09}} \\
\multicolumn{3}{c}{~}
\end{tabular}
\end{center}}\end{minipage}
}
\hspace{2em}
\subfloat[
\textbf{Low-light Enhancement}. LOL results.
\label{tab:low-light enhancement}
]{
\begin{minipage}{0.4\linewidth}{\begin{center}
\tablestyle{4pt}{1.05}
\tiny
\begin{tabular}{c|c|c}
Method & PSNR & SSIM ($\%$) \\
\shline
GLADNet~\cite{wang2018gladnet} & 19.71 & 70.30 \\
EnlightenGAN~\cite{jiang2021enlightengan} & 17.48 & 65.70 \\
KinD~\cite{zhang2019kindling} & 20.37 & 80.40 \\
MIRNet~\cite{zamir2020learning} & \underline{24.14} & 83.00 \\
night-enhancement~\cite{jin2022unsupervised} & 21.52 & 76.30 \\
HWMNet~\cite{fan2022half} & \textbf{24.24} & \underline{85.20} \\
\baseline{Restorer (Ours)} & \baseline{21.44} & \baseline{\textbf{90.53}} 
\end{tabular}
\end{center}}\end{minipage}
}
\hspace{2em}
\subfloat[
\textbf{Denoising}. SIDD denoising results.
\label{tab:denoising}
]{
\begin{minipage}{0.4\linewidth}{\begin{center}
\tablestyle{4pt}{1.05}
\tiny
\begin{tabular}{c|c|c}
Method & PSNR & SSIM ($\%$) \\
\shline
CBDNet~\cite{guo2019toward} & 30.78 & 80.10 \\
AINDNet~\cite{kim2020transfer} & 39.08 & 95.40 \\
VDN~\cite{yue2019variational} & 39.28 & 95.60 \\
SADNet~\cite{chang2020spatial} & \underline{39.46} & \underline{95.70} \\
DANet+~\cite{yue2020dual} & \textbf{39.47} & \underline{95.70} \\
DeamNet~\cite{ren2021adaptive} & \textbf{39.47} & \underline{95.70} \\
\baseline{Restorer (Ours)} & \baseline{37.75} & \baseline{\textbf{96.36}}
\end{tabular}
\end{center}}\end{minipage}
}
\hspace{2em}
\subfloat[
\textbf{Deblurring}. GoPro deblurring results.
\label{tab:deblurring}
]{
\begin{minipage}{0.4\linewidth}{\begin{center}
\tablestyle{4pt}{1.05}
\tiny
\begin{tabular}{c|c|c}
Method & PSNR & SSIM ($\%$) \\
\shline
DeblurGAN~\cite{kupyn2018deblurgan} & 28.70 & 85.80 \\
DeblurGAN-v2~\cite{kupyn2019deblurgan} & 29.55 & 93.40 \\
DBGAN~\cite{zhang2020deblurring} & 31.10 & 94.20 \\
MT-RNN~\cite{park2020multi} & {31.15} & 94.50 \\
DMPHN~\cite{zhang2019deep} & \underline{31.20} & {94.00} \\
Stripformer~\cite{tsai2022stripformer} & \textbf{33.08} & \textbf{96.20} \\
\baseline{Restorer (Ours)} & \baseline{30.92} & \baseline{\underline{95.48}}
\end{tabular}
\end{center}}\end{minipage}
}
\vspace{-1.0em}
\caption{Quantitative comparison of Restorer with several different types of image restoration expert networks.}
\label{tab2} \vspace{-2.2em}
\end{table*}

\paragraph{Textual Prompts.} Although TransWeather has shown satisfactory results in multiple severe weather image restoration tasks, the lack of a clear definition of learnable queries makes it difficult to achieve an accurate distinction between various degradation types, leading to task confusion. 
As shown in Figure~\ref{fig131}, in the deblurring task, since the blurred image also contains low-light degradation the learnable prompts-based scheme cannot accurately distinguish the degradation intended to be removed. To address this issue, we introduce textual prompts. We represent degraded types of text (e.g., ``low light", ``rain", ``blur", etc.) with the text encoder from CLIP~\cite{radford2021learning} to obtain textual prompts. 
 Specifically, we obtain the CLIP text encoder projection representation of the textual prompts $\rm{m} \in {\mathbb{R}^{\rm{1}\times\rm{D}}}$ and replicate $\rm{N}$ of them to interact with each stereo token embedding:
 \begin{equation}
   \begin{split}
         &{\rm{t}} = {\rm{repeat({\rm{CLIP(text)}})}},\\ 
  \end{split}
   \label{eq3}
 \end{equation}
 where $ \rm{CLIP} $ denotes the CLIP text encoder. $ \rm{repeat(.)} $ indicates the replicate operation. As shown in Figure~\ref{fig131}, compared with learnable queries, textual prompt with the task prior provides clear instructions for Restorer's image restoration effectively solving the task confusion issue.
In addition, Restorer can remove complex composite degenerates from an image by stacking textual prompts in the real world (see Figure~\ref{fig131}) without additional training by introducing a composite degradation dataset. 
More composite degradation restoration results are demonstrated in Appendix E.

 \paragraph{All-Axis Attention.}  As shown in the following formulation, we perform all-axis-attention modeling between stereo patches $\rm{z} \in {\mathbb{R}^{\rm{N}\times\rm{D}}}$ and the textual prompts $\rm{t} \in {\mathbb{R}^{\rm{N}\times\rm{D}}}$:
 \begin{equation}
   \begin{split}
         &{\rm{q}} = {\rm{tW_q}}, {\rm{k}} = {\rm{zW_k}}, {\rm{v}} = {\rm{zW_v}},\\ 
         &{\rm{o^{(n)}}} = \rm{\theta}(0-\rm{q^{(n)}k^{(n)\intercal}}/\sqrt{d}){\rm{v^{(n)}}},{\rm{n}}=1,...,{\rm{H}},\\
         &{\rm{o}} = [{\rm{o^{(1)}}};...;{\rm{o^{(H)}}}]{\rm{W_o}},
  \end{split}
   \label{eq1}
 \end{equation}
 where $ \rm{W_q},\rm{W_k},\rm{W_v},\rm{W_o} $ represent the mapping matrices, respectively. $ \rm{H} $ indicates the number of heads. $\rm{\theta}(.)$ is a softmax function, and $\rm{d}$ is the embedding dimension of each header to avoid gradient vanishing. 
 Unlike vanilla multi-head attention, AAA adopts negative affinity matrices. Based on these negative affinity matrices, the Restorer can assign attention weights which are not related to the degradation in all axis through latent dependencies between the textual prompts and the stereo patches, thus effectively removing the degradation information from the stereo patches in both spatial and channel (see Figure~\ref{fig4}). 
 Additionally, to further enhance the robustness of AAA to feature scales, we also established connections for AAA modules at different stages.


\begin{figure*}[!t]
    \centering
    \setlength{\abovecaptionskip}{1mm} 
    \setlength{\belowcaptionskip}{-3mm}
    \includegraphics[width=0.93\linewidth]{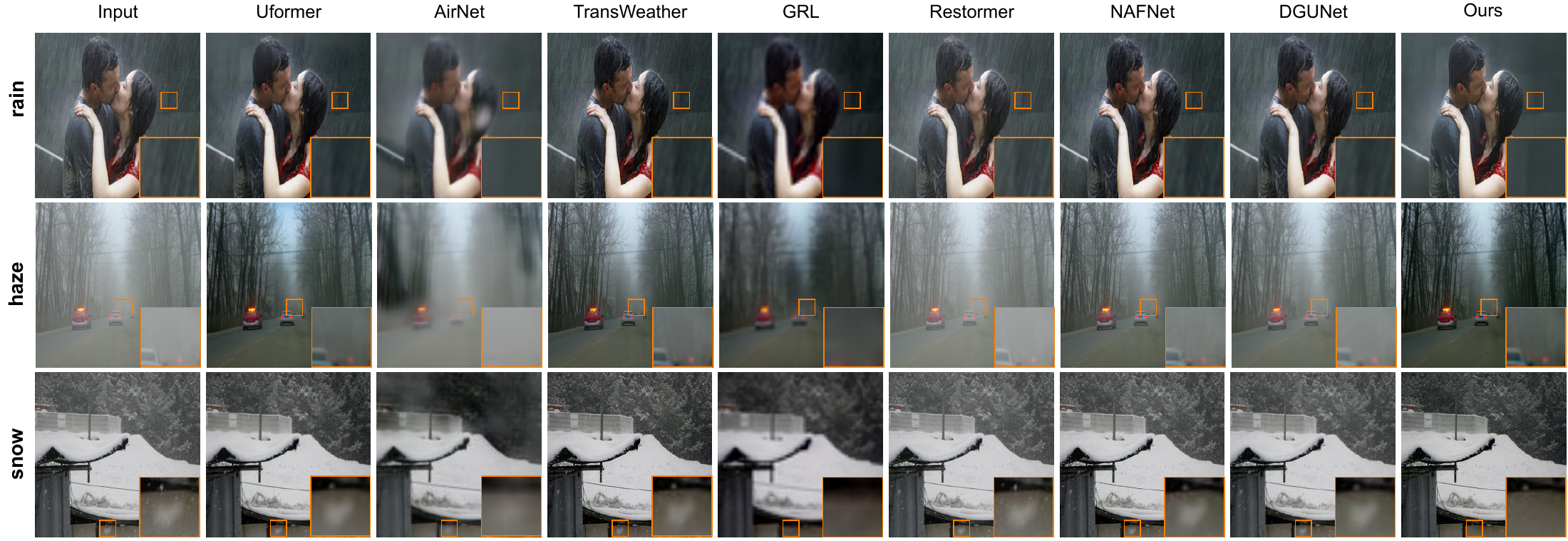}
     \caption{Sample qualitative image restoration results on the real-world rain, haze, and snow degradations.}
     \label{fig7}
   \end{figure*}

\begin{table*}[t]
\vspace{-.2em}
\centering
\subfloat[
\textbf{Ablation of Core Designs}. 
\label{tab:ablation1}
]{
\centering
\begin{minipage}{0.29\linewidth}{\begin{center}
\tablestyle{4pt}{1.05}
\resizebox{5.3cm}{!}{
\begin{tabular}{c|cccc|c}
\rowcolor{mygray}
\hline\thickhline
 & \multicolumn{4}{c|}{\text { \textbf{Module} }} & \textbf { Metric }  \\\cline{2-6}
 \rowcolor{mygray}
\multirow{-2}{*}{\textbf { Setting }} & \text { AAA } & \text{3D-DCFFN} & \text{LP} & \text{TP} &\text { PSNR/SSIM }\\
\hline\hline \text { Baseline } &  &   & & & 14.18 / 0.705\\
\text { s1 } & \checkmark && & & 25.44 / 0.902\\
\text { s2 } & \checkmark  &  \checkmark&  & & 28.26 / 0.919\\
\text { s3 } & \checkmark & \checkmark&\checkmark & &29.93 / 0.927\\
\text { s4 } & \checkmark  & \checkmark & & \checkmark & {30.28} / {0.931}\\ 
\hline\thickhline
\end{tabular}}
\end{center}}\end{minipage}
}
\hspace{2em}
\subfloat[
\textbf{Effectiveness of 3D-DCFFN}.
\label{tab:ablation2}
]{
\begin{minipage}{0.29\linewidth}{\begin{center}
\tablestyle{4pt}{1.05}
\resizebox{4.1cm}{!}{
\begin{tabular}{c|cc}
\rowcolor{mygray}
\hline\thickhline
 & \multicolumn{2}{c}{\text { \textbf{Metric} }} \\\cline{2-3}
 \rowcolor{mygray}
\multirow{-2}{*}{\textbf { Setting }} &\text { PSNR } &\text { SSIM ($\%$) }\\
\hline\hline \text { w/ vanilla FFN } &  28.14 & 92.07\\
\text { w/ DCFFN } & 28.65 & 92.38\\
\text { w/ GDFN } & 29.17 & 92.40\\
\text { w/ 3D-DCFFN } & 30.28 & 0.931\\
\hline\thickhline
\end{tabular}}
\end{center}}\end{minipage}
}
\hspace{2em}
\subfloat[
\textbf{Effectiveness of AAA}.
\label{tab:ablation3}
]{
\begin{minipage}{0.29\linewidth}{\begin{center}
\tablestyle{4pt}{1.05}
\resizebox{4.5cm}{!}{
\begin{tabular}{c|cc}
\rowcolor{mygray}
\hline\thickhline
 & \multicolumn{2}{c}{\text { \textbf{Metric} }} \\\cline{2-3}
 \rowcolor{mygray}
\multirow{-2}{*}{\textbf { Setting }} &\text { PSNR } &\text { SSIM ($\%$) }\\
\hline\hline \text { w/ spatial attention } &  25.61 & 90.33\\
\text { w/ channel attention } & 24.85 & 89.72\\
\text { w/ OSA } & 27.92 & 92.26\\
\text { w/ AAA }  & 30.28 & 0.931\\
\hline\thickhline
\end{tabular}}
\end{center}}\end{minipage}
}
\hspace{2em}
\vspace{-1em}
\caption{Ablation results for each component of Restorer. We report ablation results using desnowing experiments on CSD.}
\label{tab3} \vspace{-1.7em}
\end{table*}

 \paragraph{3D Deep Convolutional Feedforward Networks.} For stereo token embedding, the AAA module is designed with 3D Deep Convolutional Feedforward Network (3D- DCFFN) which helps Transformer bridge the deficiency of local modeling.
 Unlike other work on convolutional feedforward networks~\cite{huang2023vision,grainger2023paca,guo2022cmt}, 3D-DCFFN captures the local representation of each stereo token.
 Meanwhile, 3D-DCFFN not only considers the spatial information in each token but also pays attention to the channel information of each token at the same time.
 Specifically, we firstly map the AAA output to the hidden dimensions with a linear layer and reshape it back to stereo tokens $ \rm{E} \in {\mathbb{R}^{\rm{p}\times\rm{p}\times\rm{d}\times(\rm{n_p^2n_d})}}$. For stereo tokens, we use 3D deep convolution to extract fine-grained features in 3D space. Finally, 3D-DCFFN reshapes the stereo tokens back to 2D stereo patches and maps them back to the input dimensions by linear layer. Combining 3D-DCFFN, AAA module further improves the quality of restoration results.




 \subsection{Overall Loss}
 
 Restorer uses smooth L1-loss and perceptual loss to restore the pixels and content of the input image:
 \begin{equation}
       {L_{total}} = {L_{\rm{smoothL_1}}} + {\lambda}{L_{\rm{perceptual}}},
    \label{eq2}
  \end{equation}
 where ${\lambda}=\rm{0.04}$ controlling perceptual loss contribution.

 \section{Experiments}

 We evaluated Restorer on six different image restoration tasks. Ablation study, more experimental details, efficiency comparison, and more results can be found in the Appendix.
 
 \subsection{Experimental Setup}

 \paragraph{Datasets and metrics.} We employ PSNR and SSIM to evaluate. We unify several datasets as “\textit{mixed training set}'' for training and validate Restorer performance on official test sets, including CSD, rain1400, OTS, SIDD, LOL, GoPro, and RealBlur-R. See Appendix A for the dataset setting.

 \paragraph{Implementation details.} Restorer employs the PyTorch framework, which is trained for 250 epochs with a batch size of 26. The rest settings followed TransWeather. See Appendix B for architecture configuration and complexity.

 \subsection{Main Results}
 
 \paragraph{Comparison with unified image restoration methods.} We compare Restorer with unified image restoration methods~\cite{li2023efficient,Wang_2022_CVPR,Mou2022DGUNet,Zamir2021MPRNet,zamir2022restormer,tu2022maxim,chen2022simple} and all-in-one image restoration methods~\cite{valanarasu2022transweather,gou2023syenet,li2022all,potlapalli2023promptir,guo2024onerestore}. All algorithms are trained uniformly on “\textit{mixed training set}'' according to the official settings. 
 As shown in Figure~\ref{fig5} and Table~\ref{tab1}, our method can be successfully applied to multiple restoration tasks and exhibits competitive numerical results. 

 \paragraph{Comparison with expert networks.} We compare Restorer with expert models for each task. Some results of the compared methods are from TKL~\cite{chen2022learning}, which were all trained and tested on the corresponding single dataset. As shown in Table~\ref{tab3}, Restorer achieves comparable performance to expert networks even in an all-in-one training environment by modeling both space and channels simultaneously. 
 See Appendix F for qualitative comparisons.

 \paragraph{Real-world Tests. }We tested whether Restorer can be applied to real scenarios. Since Restorer's performance on tasks such as denoising, deblurring, and low-light enhancement are all verified on real-world datasets, we evaluate real-world severe weather restoration. Figure~\ref{fig7} shows the comparison results of severe weather restoration. Compared to baselines, Restorer is more robust in the real-world degradation.

 \subsection{Ablation Study}


 \paragraph{Effectiveness of Core Designs.} We conduct ablation studies to validate the contribution of AAA, 3D-DCFFN, and textual prompts (TP) to Restorer. The U-Net with four-stage encoder-decoder is used as the baseline. 
 Additionally, we introduced a comparison between learnable prompts (LP) and TP for Restorer. Table~\ref{tab:ablation1} shows that each design of this work contributes to Restorer. See Appendix D for more ablation.

\paragraph{Effectiveness of AAA.} To further validate the effectiveness of AAA, we also test the performance of Restorer paired with different attention mechanisms. Table~\ref{tab:ablation3} shows that AAA is more suitable for Restorer to handle multiple image restoration tasks than other attention mechanisms.

\paragraph{Effectiveness of 3D-DCFFN.} We compare the results of different feedforward networks on Restorer. Table~\ref{tab:ablation2} shows that our 3D-DCFFN better improves the quality of restored images compared to other feedforward network backbones. 

 \section{Conclusion}

 In this work, we propose Restorer to handle task confusion that exists in all-in-one image restoration. With this aim, we design an all-axis attention that has complementary advantages of spatial attention and channel attention and introduce textual prompts to solve task confusion issue. Extensive experiments have shown that Restorer has the potential to serve as a real-world image restoration application.

\clearpage
\section{Appendix}
\appendix

\section{A. Datasets}

To train Restorer which can be applied to all-in-one image restoration tasks, our “\textit{mixed training set}'' is selected from several popular image restoration datasets including CSD~\cite{hdcwnet}, rain1400~\cite{fu2017removing}, OTS~\cite{li2018benchmarking}, SIDD~\cite{abdelhamed2018high}, LOL~\cite{wei2018deep}, GoPro~\cite{nah2021ntire}, and RealBlur-R~\cite{rim2020real}. 
Detailed dataset settings are shown in Table~\ref{tab1}.

Specifically, for the desnowing task, we randomly select 5K image pairs from the CSD training set for the “\textit{mixed training set}''. In the testing phase, we use the CSD test set for testing. 
For the deraining task, we randomly select 5,000 image pairs from the rain1400 training set for the “\textit{mixed training set}''.
We test the model using the rain1400 test set, containing 1,400 image pairs.
For the defogging task, the “\textit{mixed training set}'' contains 5,000 image pairs, which are randomly selected from the OTS dataset following the settings of Chen et al.~\cite{chen2022learning}. We applied SOTS, which is a test set of RESIDE defogging dataset~\cite{reside}.
For denoising, the “\textit{mixed training set}'' includes 5,000 image pairs randomly selected from SIDD dataset, and we utilize the SIDD test set for testing.
For the low-light enhancement task, we employ the LOL dataset, containing 485 training image pairs and 15 test pairs. 
Finally, for deblurring, the “\textit{mixed training set}'' includes 2,103 GoPro and 3,758 RealBlur-R training pairs. For the testing of deblurring task, we use the test sets of GoPro and RealBlur-R respectively. 

\begin{table}[h]
\centering
\setlength{\abovecaptionskip}{0.1cm} 
\setlength{\belowcaptionskip}{-0.7cm}
\scriptsize
\setlength{\tabcolsep}{4pt}
\renewcommand{\arraystretch}{1.1}
\begin{tabular}{l|l|rr|l}
 \textbf{Task}  & \textbf{Dataset}  & \textbf{\#Train} & \textbf{\#Test} & \textbf{Test Dubname} \\
 \toprule
 \multirow{1}{*}{\footnotesize\text{Desnowing}} & CSD & 5,000 & 2,000 & CSD \\
 \hline
 \multirow{1}{*}{\footnotesize\text{Deraining}} & 
 rain1400 & 5,000 & 1,400 & rain1400 \\
 \hline
 \multirow{1}{*}{\footnotesize\text{Dehazing}} & 
 OTS & 5,000 & 500 & SOTS \\
 \hline
 \multirow{1}{*}{\footnotesize\text{Denoising}} 
 & SIDD & 5,000 & 1,280 & SIDD \\
 \hline
 \multirow{2}{*}{\footnotesize\text{Debluring}} 
 & GoPro & 2,103 & 1,111 & GoPro \\
 & RealBlur-R & 3,758 & 980 & RealBlur-R \\
 \hline
{\footnotesize\text{Enhancement}} 
 & LOL & 485 & 15 & LOL \\
\bottomrule
\end{tabular}
\caption{Dataset configuration on six image restoration tasks.}
\label{tab1}
\end{table}

\section{B. Computational Complexity}

\begin{table}[h]
\centering
\scriptsize
\setlength{\tabcolsep}{4pt}
\renewcommand{\arraystretch}{1.1}
\begin{tabular}{c|l|l|c}
\hline\thickhline
\multicolumn{4}{c}{ \text{Architecture} }\\\cline{1-4}
Stage  & Input shape & Output Shape & Layers  \\
\toprule
\multirow{2}{*}{\text { Stage1 }}  & \multirow{2}{*}{$256^2\times 3$} &  \multirow{2}{*}{$128^2\times 128$}&$ \text{N}_1=3,\text{p}=16,$  \\
&&&$ \text{d}=16,\text{D}=512,\text{H}=2$\\
\midrule
\multirow{2}{*}{\text { Stage2 }}  & \multirow{2}{*}{$128^2\times 128$} &  \multirow{2}{*}{$64^2\times 128$} & $\text{N}_2=4,\text{p}=8,$ \\
&&&$ \text{d}=16,\text{D}=512,\text{H}=2$\\
\midrule
\multirow{2}{*}{\text { Stage3 }}  & \multirow{2}{*}{$64^2\times 128$} &  \multirow{2}{*}{$32^2\times 256$} & $\text{N}_3=6,\text{p}=4,$ \\
&&&$ \text{d}=32,\text{D}=512,\text{H}=4$\\
\midrule
\multirow{2}{*}{\text { Stage4 }}  & \multirow{2}{*}{$32^2\times 256$} &  \multirow{2}{*}{$16^2\times 512$} & $\text{N}_4=3,\text{p}=4,$ \\
&&&$ \text{d}=16,\text{D}=512,\text{H}=8$\\
\bottomrule
\end{tabular}
\caption{Detailed architectural configurations for the different stages of Restorer.}
\label{tab2}
\end{table}

\begin{table}[h]
\centering
\setlength{\abovecaptionskip}{0.3cm} 
\setlength{\belowcaptionskip}{-0.2cm}
\footnotesize
\setlength{\tabcolsep}{2pt}
\begin{tabular}{l|l|l|c|r}
\textbf{Task} & \textbf{Dataset} & \textbf{Model} & \textbf{PSNR} &  \textbf{FLOPs} \\ 
\toprule
\multirow{4}{*}{Denoise} & \multirow{4}{*}{SIDD} & Restormer & 25.30 & 141G \\
& &  MAXIM & 27.70 & 216G \\
& & \textbf{Ours} & 37.75 & 147G \\
\hline
\multirow{4}{*}{Deblur} & \multirow{4}{*}{GoPro} 
& Restormer & 26.22 & 141G \\
& &  MAXIM & 25.78 & 216G \\
& & \textbf{Ours} & 30.92 & 147G \\
\hline
\multirow{4}{*}{Deblur} & \multirow{4}{*}{RealBlur-R} 
& Restormer & 34.42 & 141G \\
& &  MAXIM & 32.52 & 216G \\
& & \textbf{Ours} & 43.10 & 147G \\
\hline
\multirow{4}{*}{Derain} & \multirow{4}{*}{rain1400} & Restormer & 24.38 & 141G \\
& &  MAXIM & 32.63 & 216G \\
& & \textbf{Ours} & 34.21 & 147G \\
\hline
\multirow{4}{*}{Dehaze} & \multirow{4}{*}{SOTS} 
& Restormer & 15.94 & 141G \\
& &  MAXIM & 32.20 & 216G \\
& & \textbf{Ours} & 34.46 & 147G \\
\hline
\multirow{4}{*}{Enhance} & \multirow{4}{*}{LOL} 
& Restormer & 7.79 & 141G \\
& &  MAXIM & 13.48 & 216G \\
& & \textbf{Ours} & 21.44 & 147G \\

\bottomrule
\end{tabular}
\caption{Model performance vs. complexity comparison of our model with SOTA baselines.}
\label{tab3}
\end{table}

\begin{figure*}[!t]
    \centering
    \setlength{\abovecaptionskip}{1mm} 
    \setlength{\belowcaptionskip}{-1mm}
    \includegraphics[width=1\linewidth]{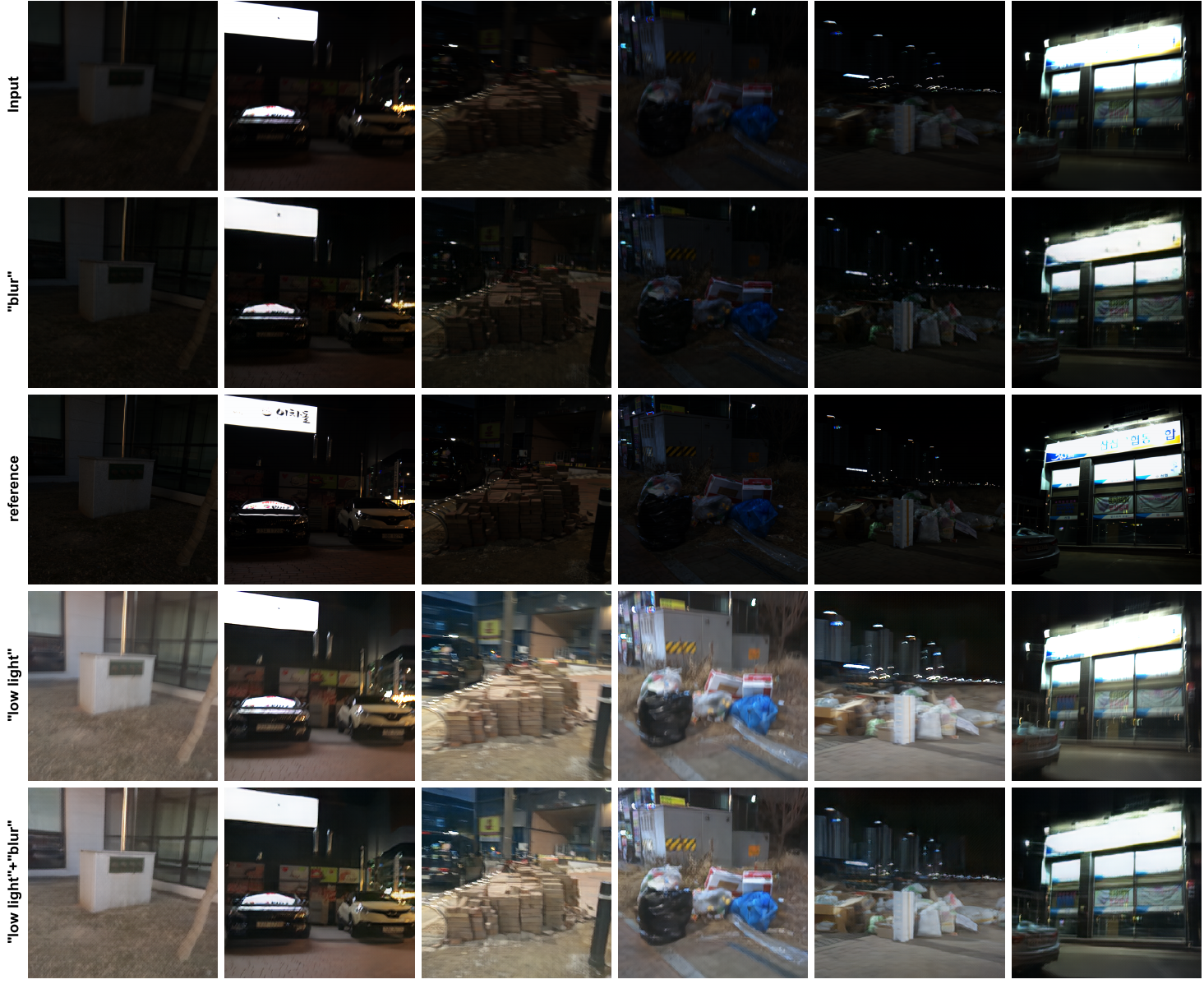}
     \caption{Restorer's image restoration results at different textual prompts on RealBlur-R.}
     \label{Fig3}
   \end{figure*}

\begin{figure*}[!t]
    \centering
    \setlength{\abovecaptionskip}{1mm} 
    \setlength{\belowcaptionskip}{-1mm}
    \includegraphics[width=1\linewidth]{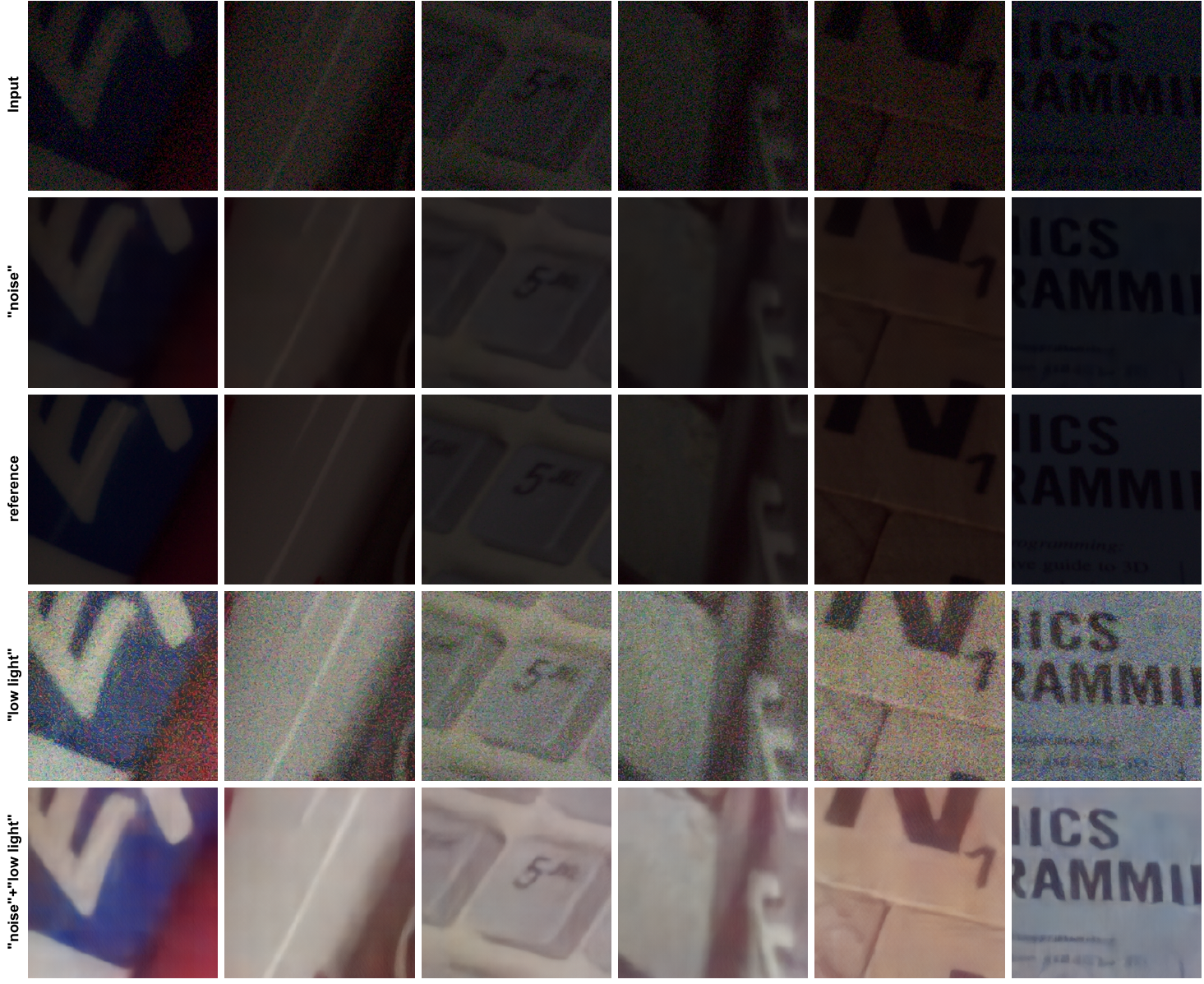}
     \caption{Restorer's image restoration results at different textual prompts on SIDD.}
     \label{Fig3(1)}
   \end{figure*}

The detailed specifications of the Restorer architecture configuration are shown in Table~\ref{tab2}. We provide the input shapes and output shapes for each stage of the Restorer encoder for better understanding. 
We reduce the computational overhead of AAA module by adjusting the spatial dimension $\text{p}$ and the channel dimension $\text{d}$ of the stereo token embedding at each stage. Also for this purpose, we unify the embedding dimension $\text{D}$ of each stage to 512 dimensions. See Table~\ref{tab3} for a comparison of performance and complexity for our method with Restormer~\cite{zamir2022restormer} and MAXIM~\cite{tu2022maxim}.
Moreover, we show the computational complexity of each component in Restorer in Table~\ref{tab31}. It can be noticed that our AAA module and 3D-DCFFN lead to a relatively small computation cost.

\begin{table}[!h]
\centering
\setlength{\abovecaptionskip}{0.1cm} 
\setlength{\belowcaptionskip}{-0.2cm}
\resizebox{6cm}{!}{
\begin{tabular}{c|ccc|c}
\rowcolor{mygray}
\hline\thickhline
& \multicolumn{3}{c|}{\text { \textbf{Module} }} & \\\cline{2-4}
\rowcolor{mygray}
\multirow{-2}{*}{\textbf { Setting }} & \text { AAA } & \text{3D-DCFFN} & \text{TP}&\multirow{-2}{*}{\textbf { FLOPs }} \\
\hline\hline \text { Baseline } &  &   & & 147G\\
\text { s1 } &  & \checkmark& \checkmark& 121G\\
\text { s2 } &\checkmark &  & \checkmark & 144G\\
\text { s3 } & \checkmark & \checkmark & &147G\\
\hline\thickhline
\end{tabular}}
\caption{Ablation of Core Designs. TP represents textual prompts.}
\label{tab31}
\end{table}
\vspace{-0.5cm}

\section{C. Experimental Details}

Restorer mainly follows the settings of~\cite{valanarasu2022transweather}. 
Apart from the training settings described in the main paper, Restorer takes image with resolution of 256x256 as input, and performs normalization for the input data. 
In addition, for training after 60 epochs, the learning rate decays by half every 50 epochs. We train the Restorer using two NVIDIA RTX 3090.

\section{D. Ablation Study}

\paragraph{AAA Connection.} We remove the connection between the AAA modules. As shown in Table~\ref{tab:ablation4}, after removing the connection between the AAA modules, Restorer's performance on the desnowing task decreases to some extent, suggesting the effectiveness of this design.

\paragraph{Encoder and Decoder.} For the ablation of Restorer's encoder and decoder, we apply the encoders and decoders in Restormer~\cite{zamir2022restormer}, which employ the Transformer as a base module, to replace the corresponding components in Restorer, respectively. 
In Table~\ref{tab:ablation3} and Table~\ref{tab:ablation2}, we observe that either of the ablations for the encoder and decoder in Restorer resulted in performance degradation in Restorer's metrics.

\begin{table*}[t]
\vspace{-.2em}
\centering
\subfloat[
Ablation results for AAA connections.
\label{tab:ablation4}
]{
\centering
\begin{minipage}{0.359\linewidth}{\begin{center}
\tablestyle{4pt}{1.05}
\resizebox{5cm}{!}{
\begin{tabular}{c|cc}
\rowcolor{mygray}
\hline\thickhline
& \multicolumn{2}{c}{\text { \textbf{Metric} }}  \\\cline{2-3}
\rowcolor{mygray}
\multirow{-2}{*}{\textbf { Method }} &\text {PSNR}& \text {SSIM}\\
\hline\hline \text { w/o AAA connection } & 29.55 & 0.924 \\
\text { w/ AAA connection } & 30.28 & 0.931 \\

\hline\thickhline
\end{tabular}}
\end{center}}\end{minipage}
}
\hspace{2em}
\subfloat[
Ablation results for the decoder in Restorer.
\label{tab:ablation3}
]{
\centering
\begin{minipage}{0.4\linewidth}{\begin{center}
\tablestyle{4pt}{1.05}
\resizebox{5cm}{!}{
\begin{tabular}{c|cc}
\rowcolor{mygray}
\hline\thickhline
& \multicolumn{2}{c}{\text { \textbf{Metric} }}  \\\cline{2-3}
\rowcolor{mygray}
\multirow{-2}{*}{\textbf { Method }} &\text {PSNR}& \text {SSIM}\\
\hline\hline 
\text { w/ Restormer encoder } & 29.97 & 0.926 \\
\text { w/ Restorer encoder } & 30.28 & 0.931 \\

\hline\thickhline
\end{tabular}}
\end{center}}\end{minipage}
}
\hspace{2em}
\subfloat[
Ablation results for the encoder in Restorer.
\label{tab:ablation2}
]{
\centering
\begin{minipage}{0.39\linewidth}{\begin{center}
\tablestyle{4pt}{1.05}
\resizebox{4.8cm}{!}{
\begin{tabular}{c|cc}
\rowcolor{mygray}
\hline\thickhline
& \multicolumn{2}{c}{\text { \textbf{Metric} }}  \\\cline{2-3}
\rowcolor{mygray}
\multirow{-2}{*}{\textbf { Method }} &\text {PSNR}& \text {SSIM}\\
\hline\hline 
\text { w/ Restormer decoder } & 30.06 & 0.929 \\
\text { w/ Restorer decoder } & 30.28 & 0.931 \\

\hline\thickhline
\end{tabular}}
\end{center}}\end{minipage}
}
\hspace{2em}
\subfloat[
Effectiveness of negative affinity matrices in AAA.
\label{tab:ablation1}
]{
\centering
\begin{minipage}{0.4\linewidth}{\begin{center}
\tablestyle{4pt}{1.05}
\resizebox{5cm}{!}{
\begin{tabular}{c|cc}
\rowcolor{mygray}
\hline\thickhline
& \multicolumn{2}{c}{\text { \textbf{Metric} }}  \\\cline{2-3}
\rowcolor{mygray}
\multirow{-2}{*}{\textbf { Method }} &\text {PSNR}& \text {SSIM}\\
\hline\hline \text { w/ vanilla affinity matrices } & 30.11 & 0.927 \\
\text { w/ projection affinity matrices } & 30.07 & 0.927 \\
\text { w/ negative affinity matrices } & 30.28 & 0.931 \\
\hline\thickhline
\end{tabular}}
\end{center}}\end{minipage}
}
\hspace{2em}
\vspace{-.1em}
\caption{Ablation results for each component of Restorer. We report ablation results using desnowing experiments on CSD.}
\label{taba} \vspace{-.5em}
\end{table*}

\paragraph{Negative Affinity Matrices.} We test the effectiveness of negative affinity matrices coupled with textual prompts in all-axis attention in this section. We compare the negative affinity matrices with two variants, the vanilla affinity matrices, and the projection affinity matrices. As shown in Table~\ref{tab:ablation1}, compared to other types of affinity matrices, Restorer with negative affinity matrices can achieve more accurate image restoration.

\section{E. Composite Degradation Restoration}

In this section, we provide more visual results of Restorer employing different textual prompts for the same degraded image. We also demonstrate the effectiveness of Restorer in restoration of complex composite degraded images by iterating over the restoration results with several textual prompts.

As shown in Figure~\ref{Fig3}, Restorer successfully performs different image restoration tasks on the RealBlur-R dataset following different textual prompts. 
Meanwhile, by iterating textual prompts, Restorer successfully restores input images with complex composite degradation. 
Furthermore, we also show the performance of Restorer on SIDD based on different textual prompts in Figure~\ref{Fig3(1)}. 
Restorer strictly follows the textual prompts to perform the appropriate image restoration tasks. Moreover, by iterating textual prompts Restorer removes the complex degradation of noise and low-light composites. 
We hope that this finding will be useful for future complex composite degraded image restoration tasks.

\begin{figure*}[!t]
    \centering
    \setlength{\abovecaptionskip}{1mm} 
    \setlength{\belowcaptionskip}{-6mm}
    \includegraphics[width=1\linewidth]{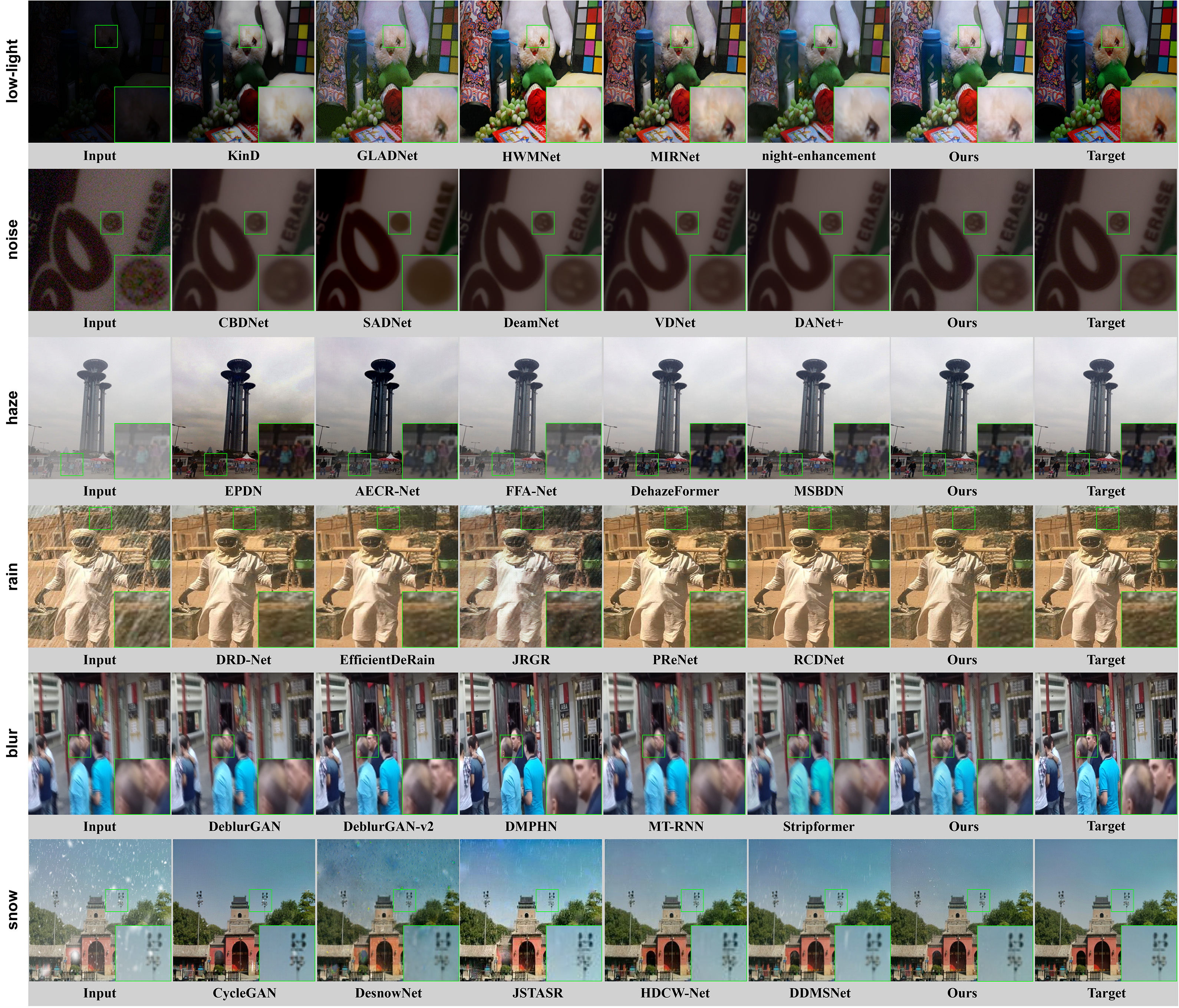}
     \caption{Qualitative comparison results of Restorer with expert networks on individual image restoration tasks.}
     \label{figx}
   \end{figure*}

\section{F. More Experiments}

\paragraph{Comparison with Expert Networks. }In Figure~\ref{figx}, we compare the all-in-one trained Restorer with expert models on several image restoration tasks. These results suggest that Restorer accurately performs the appropriate image restoration tasks according to the textual prompts without task confusion issues. 
At the same time, Restorer effectively removes most of degradation through AAA's synchronized modeling of spatial and channel dimensions.
In contrast, some expert networks still demonstrate residual degradation.

\begin{figure*}[!t]
    \centering
    \setlength{\abovecaptionskip}{1mm} 
    \setlength{\belowcaptionskip}{-1mm}
    \includegraphics[width=0.93\linewidth]{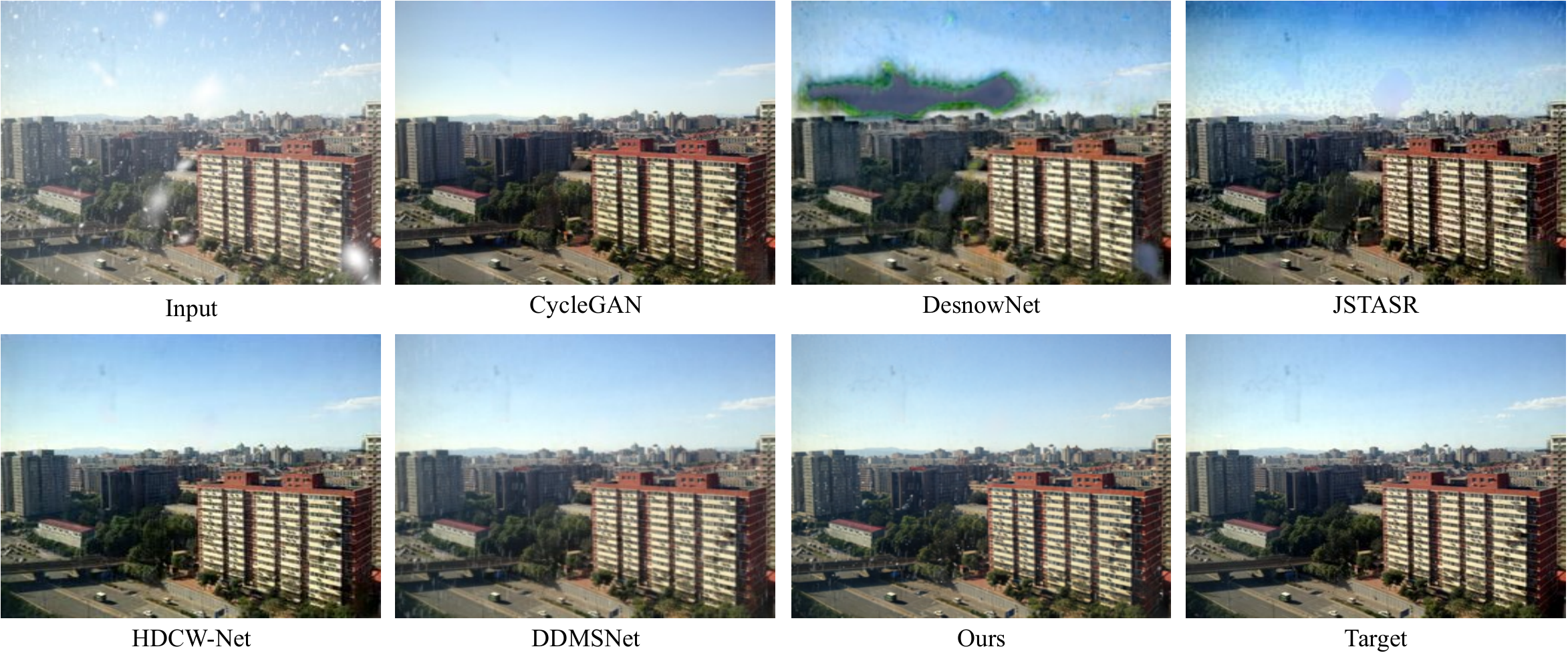}
     \caption{Visual comparison results between Restorer and desnowing baselines on CSD.}
     \label{Fig5}
   \end{figure*}

\begin{figure*}[!t]
    \centering
    \setlength{\abovecaptionskip}{1mm} 
    \setlength{\belowcaptionskip}{-1mm}
    \includegraphics[width=0.93\linewidth]{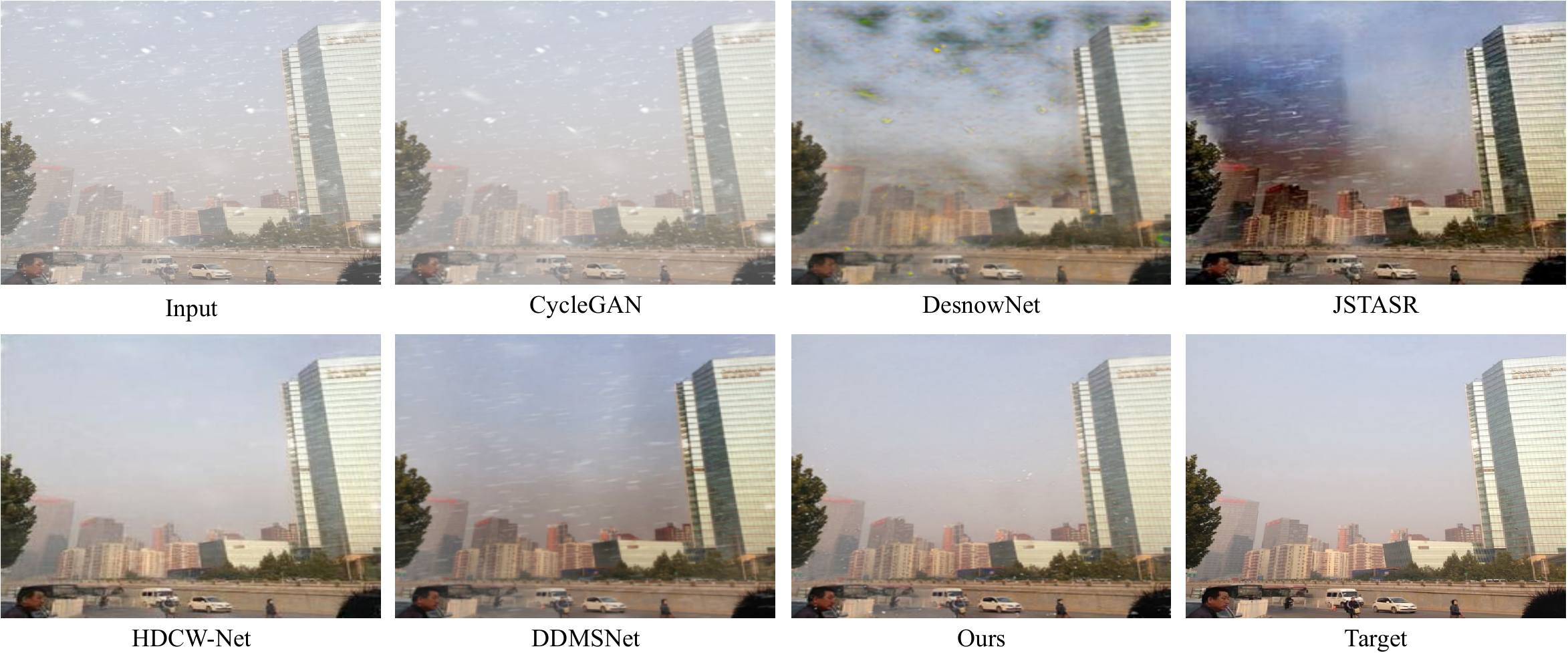}
     \caption{Visual comparison results between Restorer and desnowing baselines on CSD.}
     \label{Fig5(1)}
   \end{figure*}

\paragraph{Desnowing.} As shown in Figure~\ref{Fig5} and Figure~\ref{Fig5(1)}, our method successfully removes snow particles of different sizes from the input image without artifacts and chromatic aberration problems compared to multiple desnowing expert networks, which were trained and tested on CSD dataset.

\begin{figure*}[!t]
    \centering
    \setlength{\abovecaptionskip}{1mm} 
    \setlength{\belowcaptionskip}{-1mm}
    \includegraphics[width=0.93\linewidth]{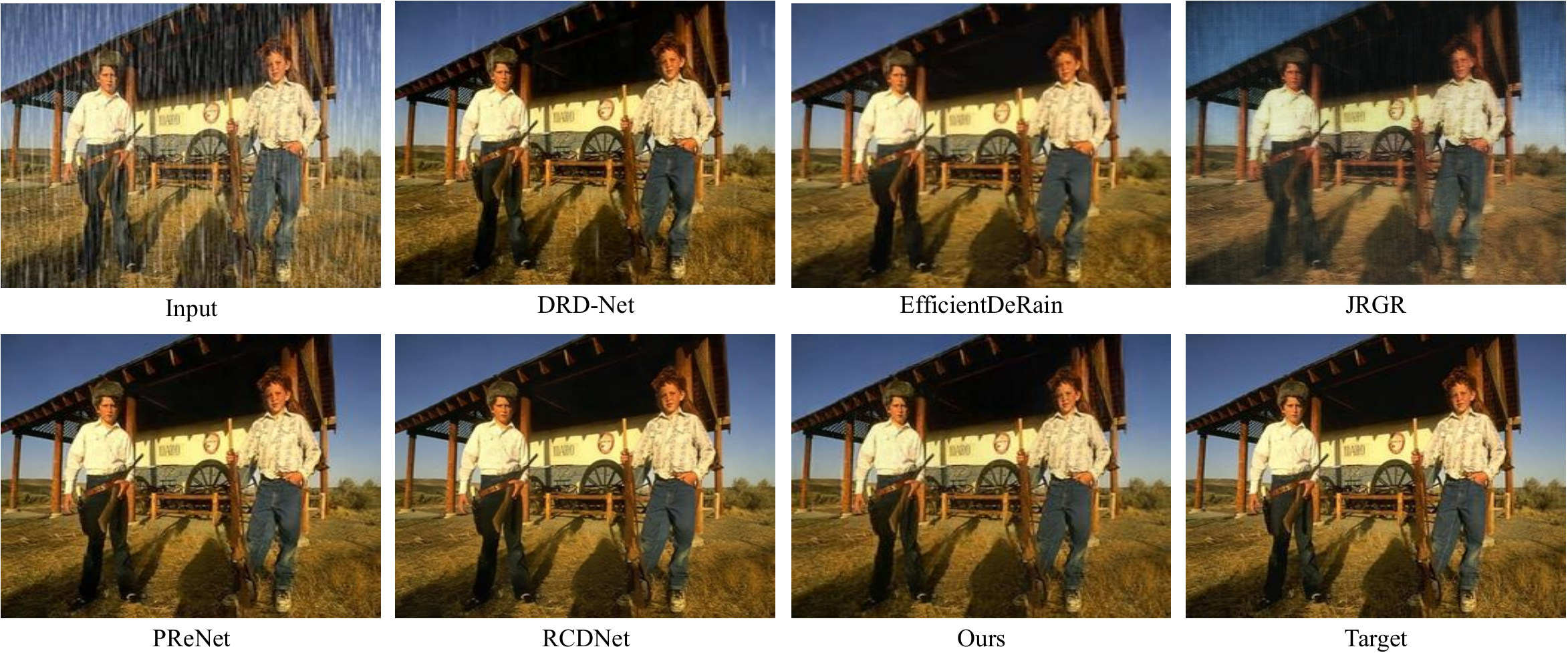}
     \caption{Visual comparison results between Restorer and deraining baselines on rain1400.}
     \label{Fig6}
   \end{figure*}
\begin{figure*}[!t]
    \centering
    \setlength{\abovecaptionskip}{1mm} 
    \setlength{\belowcaptionskip}{-1mm}
    \includegraphics[width=0.93\linewidth]{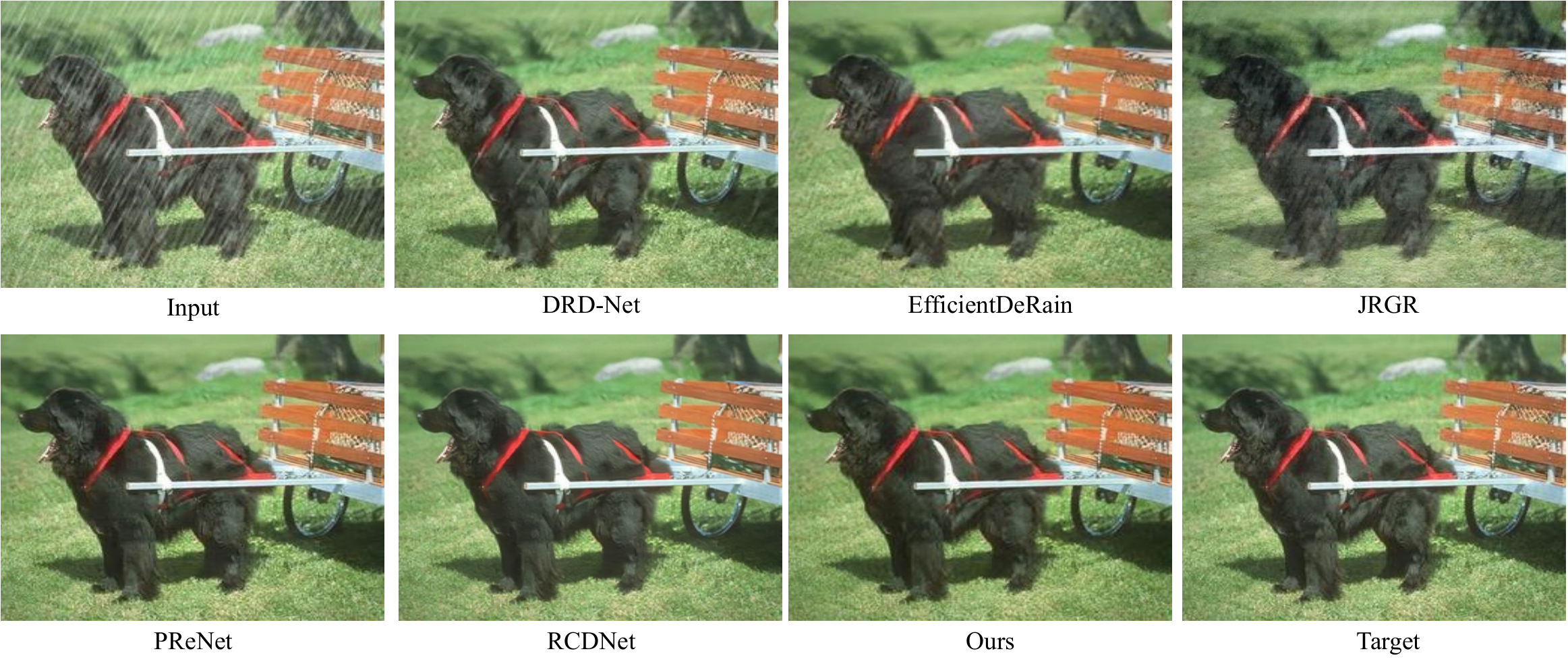}
     \caption{Visual comparison results between Restorer and deraining baselines on rain1400.}
     \label{Fig6(1)}
   \end{figure*}

\paragraph{Deraining.} Figure~\ref{Fig6} and Figure~\ref{Fig6(1)} shows the visual comparison results between Restorer and the current deraining expert models on rain1400. It can be observed that our method effectively removes the rain from the input image and successfully obtains a high-quality restored image.

\begin{figure*}[!t]
    \centering
    \setlength{\abovecaptionskip}{1mm} 
    \setlength{\belowcaptionskip}{-1mm}
    \includegraphics[width=0.93\linewidth]{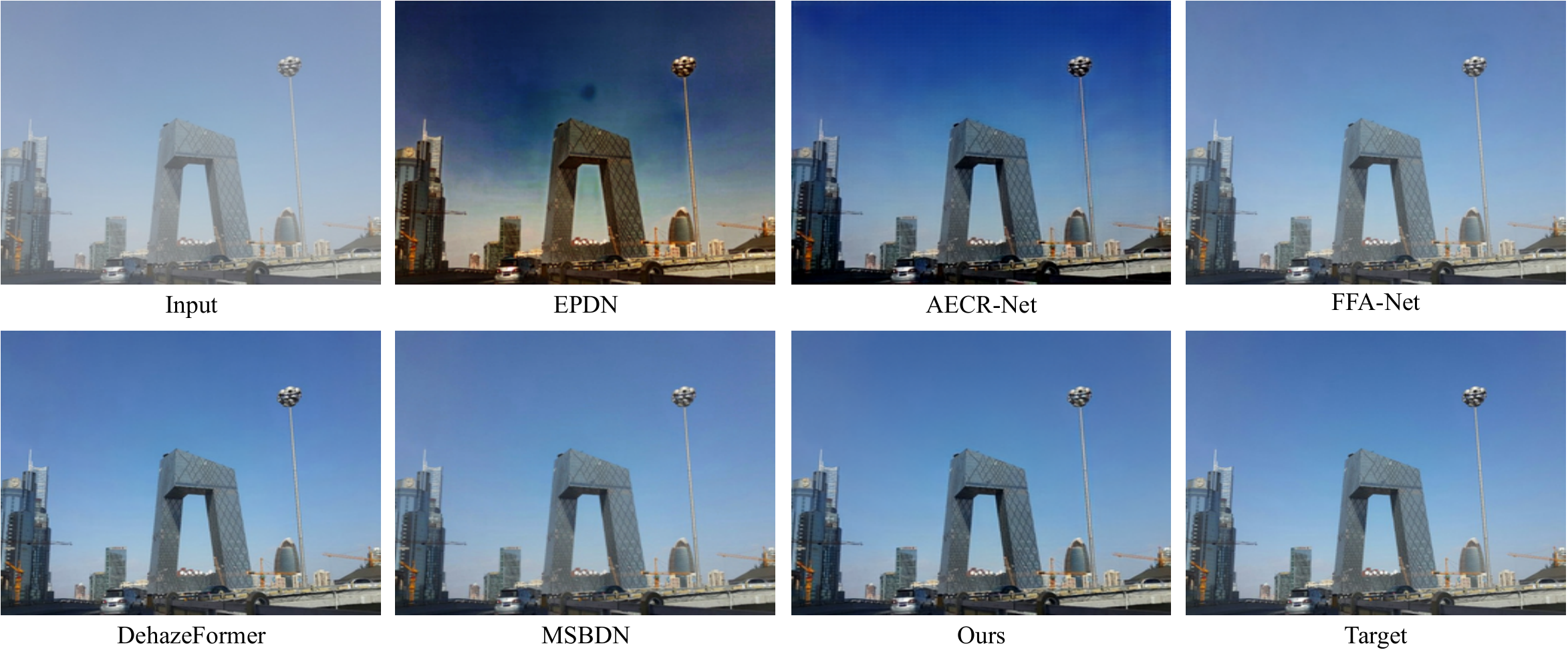}
     \caption{Visual comparison results between Restorer and defogging baselines on SOTS.}
     \label{Fig7}
   \end{figure*}

\begin{figure*}[!t]
    \centering
    \setlength{\abovecaptionskip}{1mm} 
    \setlength{\belowcaptionskip}{-1mm}
    \includegraphics[width=0.93\linewidth]{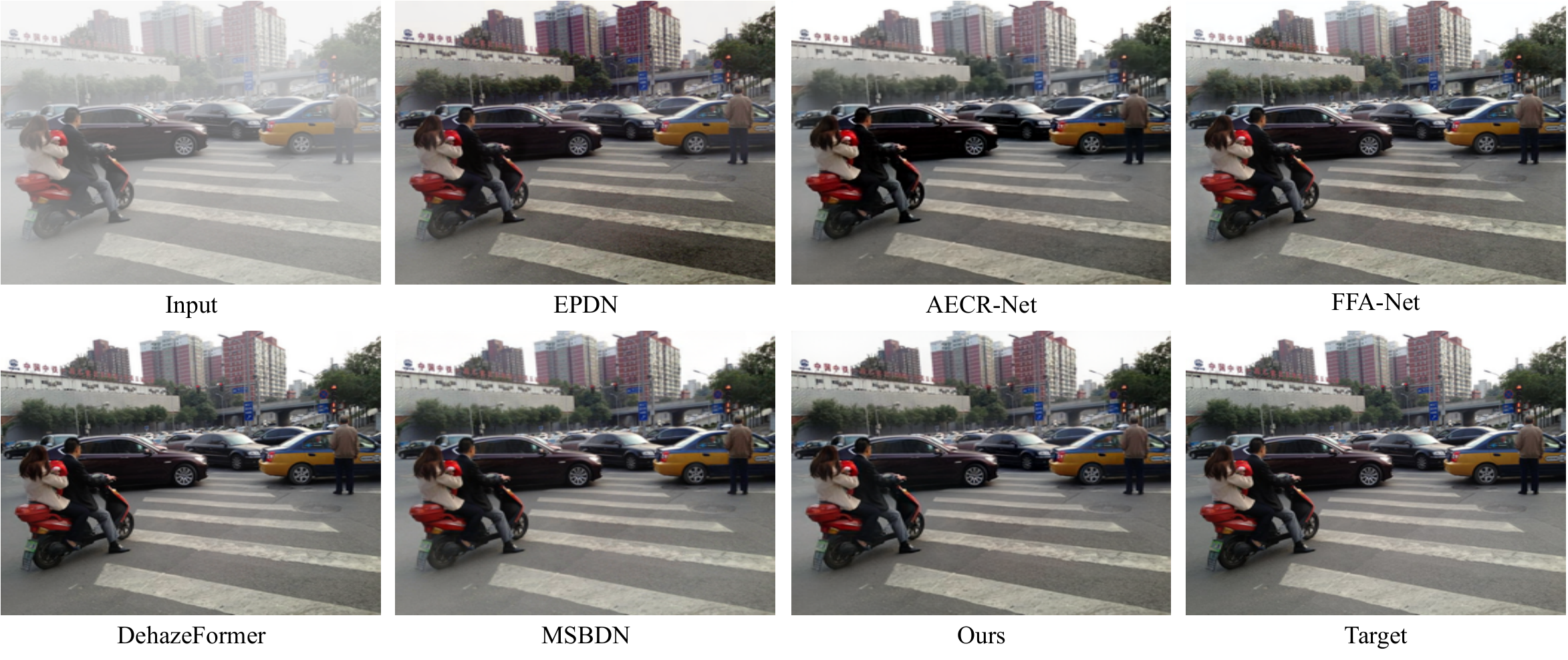}
     \caption{Visual comparison results between Restorer and defogging baselines on SOTS.}
     \label{Fig7(1)}
   \end{figure*}

\paragraph{Defogging.} We provide a defogging visual comparison in Figure~\ref{Fig7} and Fig~\ref{Fig7(1)}. 
Although all the comparison algorithms successfully remove the degradation, EPDN~\cite{qu2019enhanced}, AECR-Net~\cite{wu2021contrastive}, and FFA-Net~\cite{qin2020ffa} produce severe chromatic aberration affecting the quality of the restored images. On the other hand, DehazeFormer~\cite{song2023vision} and MSBDN~\cite{dong2020multi} demonstrate some fog residuals in the “\textit{house area}''. In comparison, Restorer demonstrates a more obvious defogging effect and more satisfactory image restoration results.

\begin{figure*}[!t]
    \centering
    \setlength{\abovecaptionskip}{1mm} 
    \setlength{\belowcaptionskip}{-1mm}
    \includegraphics[width=0.93\linewidth]{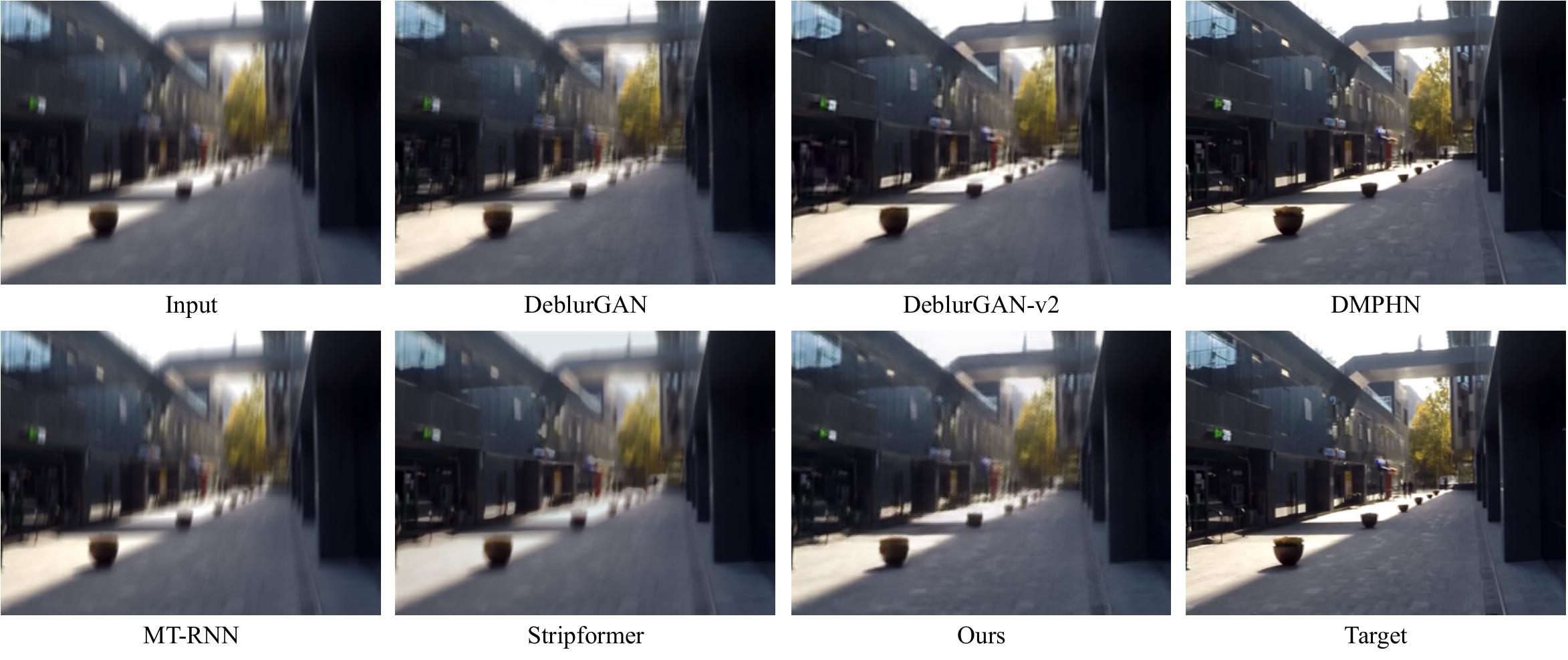}
     \caption{Visual comparison results between Restorer and deblurring baselines on GoPro.}
     \label{fig8}
   \end{figure*}

\begin{figure*}[!t]
    \centering
    \setlength{\abovecaptionskip}{1mm} 
    \setlength{\belowcaptionskip}{-1mm}
    \includegraphics[width=0.93\linewidth]{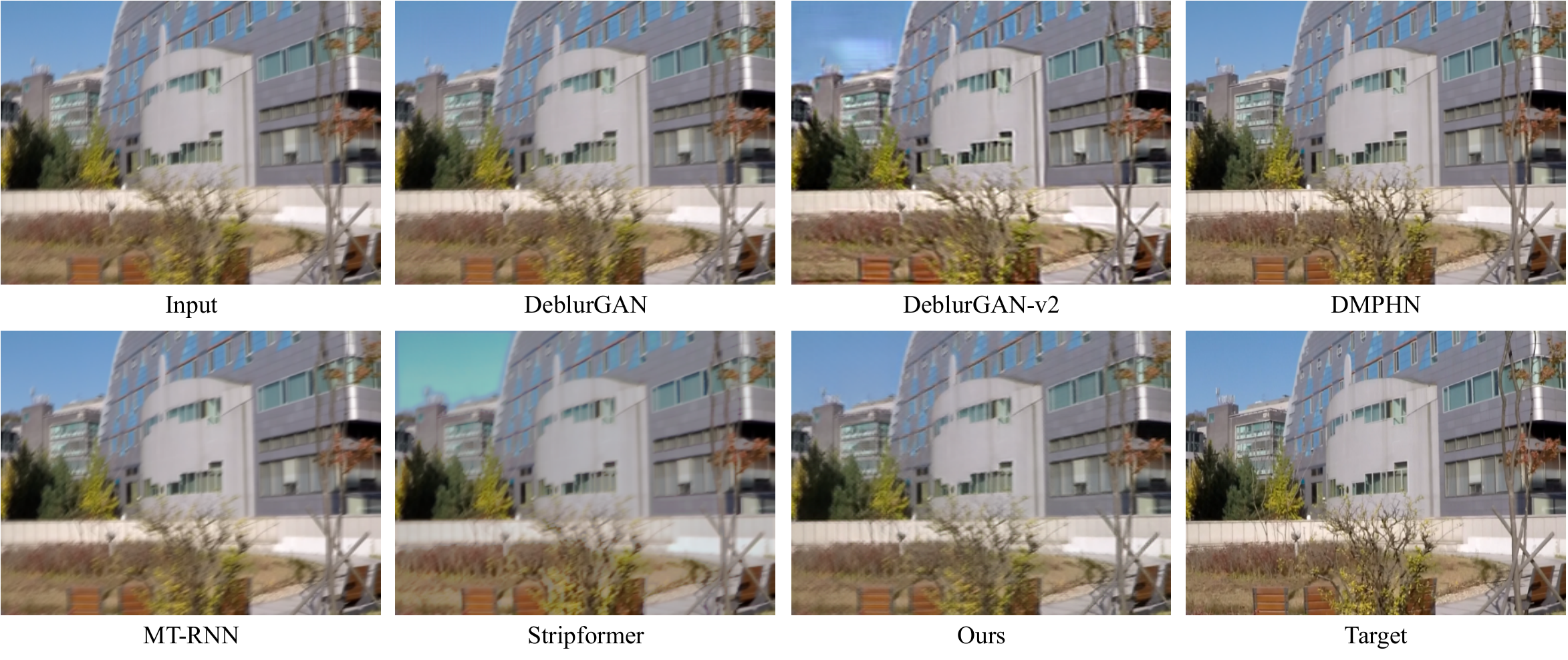}
     \caption{Visual comparison results between Restorer and deblurring baselines on GoPro.}
     \label{fig8(1)}
   \end{figure*}

\begin{figure*}[!t]
    \centering
    \setlength{\abovecaptionskip}{1mm} 
    \setlength{\belowcaptionskip}{-1mm}
    \includegraphics[width=0.93\linewidth]{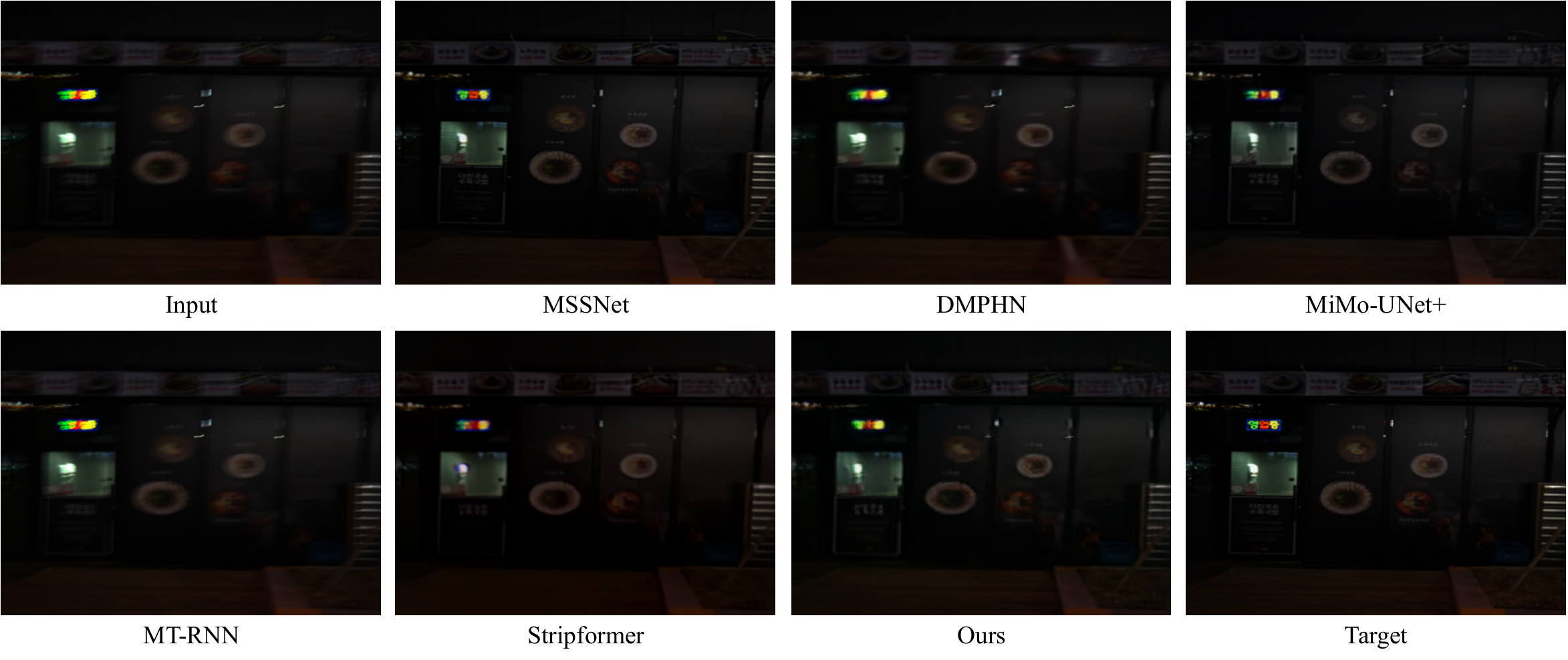}
     \caption{Visual comparison results between Restorer and deblurring baselines on RealBlur-R.}
     \label{Fig9}
   \end{figure*}

\begin{figure*}[!t]
    \centering
    \setlength{\abovecaptionskip}{1mm} 
    \setlength{\belowcaptionskip}{-1mm}
    \includegraphics[width=0.93\linewidth]{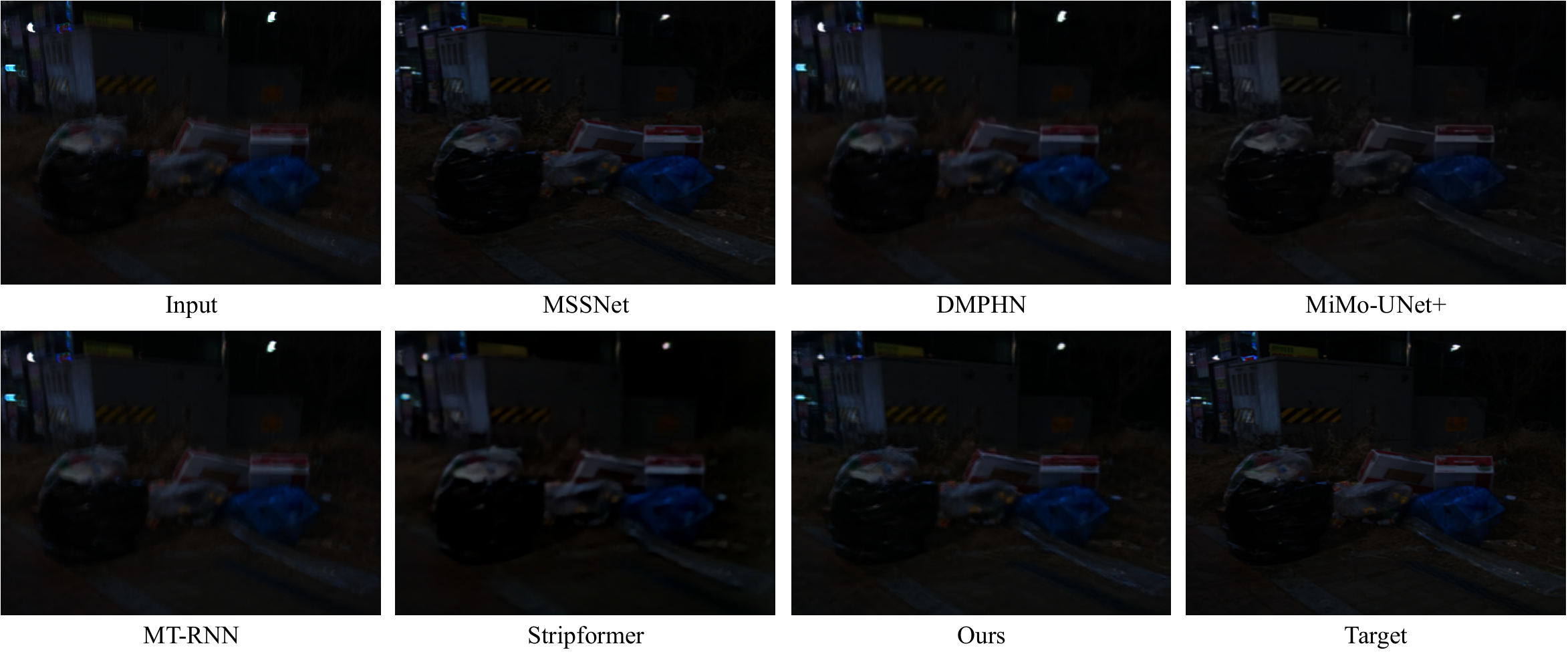}
     \caption{Visual comparison results between Restorer and deblurring baselines on RealBlur-R.}
     \label{fig9(1)}
   \end{figure*}

\paragraph{Deblurring.} The visualizations on GoPro~\cite{nah2021ntire} and RealBlur-R~\cite{rim2020real} are shown in Figure~\ref{fig8}, Figure~\ref{fig8(1)}, Figure~\ref{Fig9}, and Figure~\ref{fig9(1)}, respectively. Our model achieves competitive results on both real-world deblurring benchmarks, suggesting the robustness of Restorer for multiple types of blur.

\begin{figure*}[!t]
    \centering
    \setlength{\abovecaptionskip}{1mm} 
    \setlength{\belowcaptionskip}{-1mm}
    \includegraphics[width=0.93\linewidth]{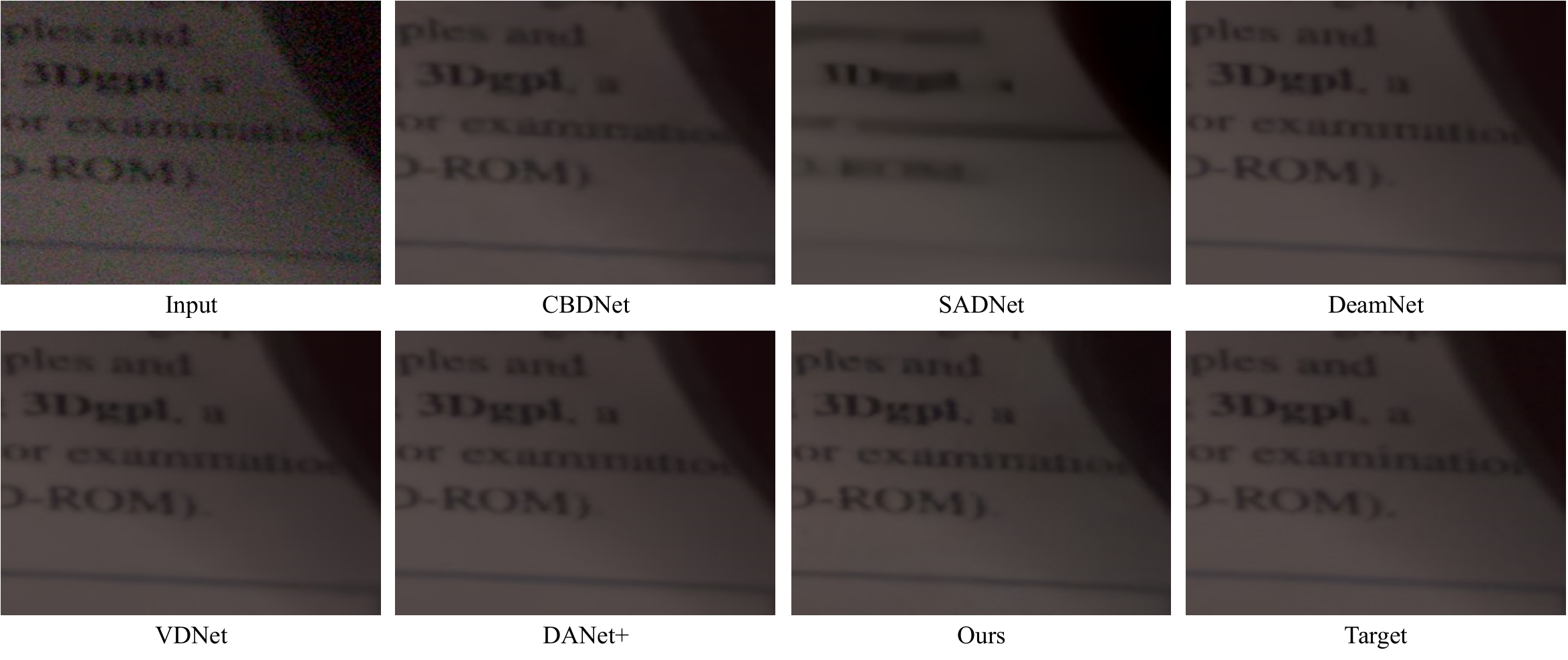}
     \caption{Visual comparison results between Restorer and denoising baselines on SIDD.}
     \label{Fig10}
   \end{figure*}
\begin{figure*}[!t]
    \centering
    \setlength{\abovecaptionskip}{1mm} 
    \setlength{\belowcaptionskip}{-1mm}
    \includegraphics[width=0.93\linewidth]{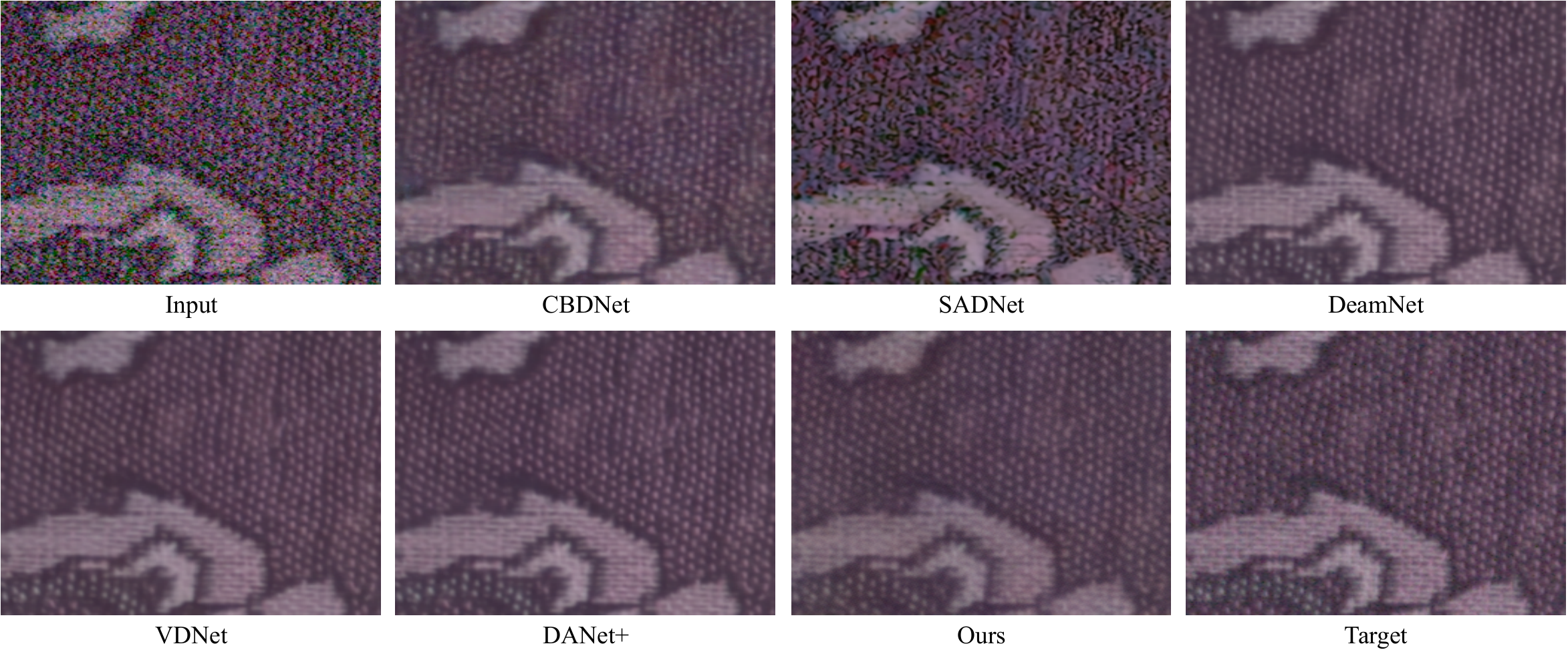}
     \caption{Visual comparison results between Restorer and denoising baselines on SIDD.}
     \label{Fig10(1)}
   \end{figure*}

\paragraph{Denoising.} The qualitative comparison results of the denoising task for the SIDD dataset are shown in Figure~\ref{Fig10} and Figure~\ref{Fig10(1)}. Both our method and the comparison baselines successfully remove the noise from the input image. But Restorer still provides clearer restored results than most of baselines.

\begin{figure*}[!t]
    \centering
    \setlength{\abovecaptionskip}{1mm} 
    \setlength{\belowcaptionskip}{-1mm}
    \includegraphics[width=0.93\linewidth]{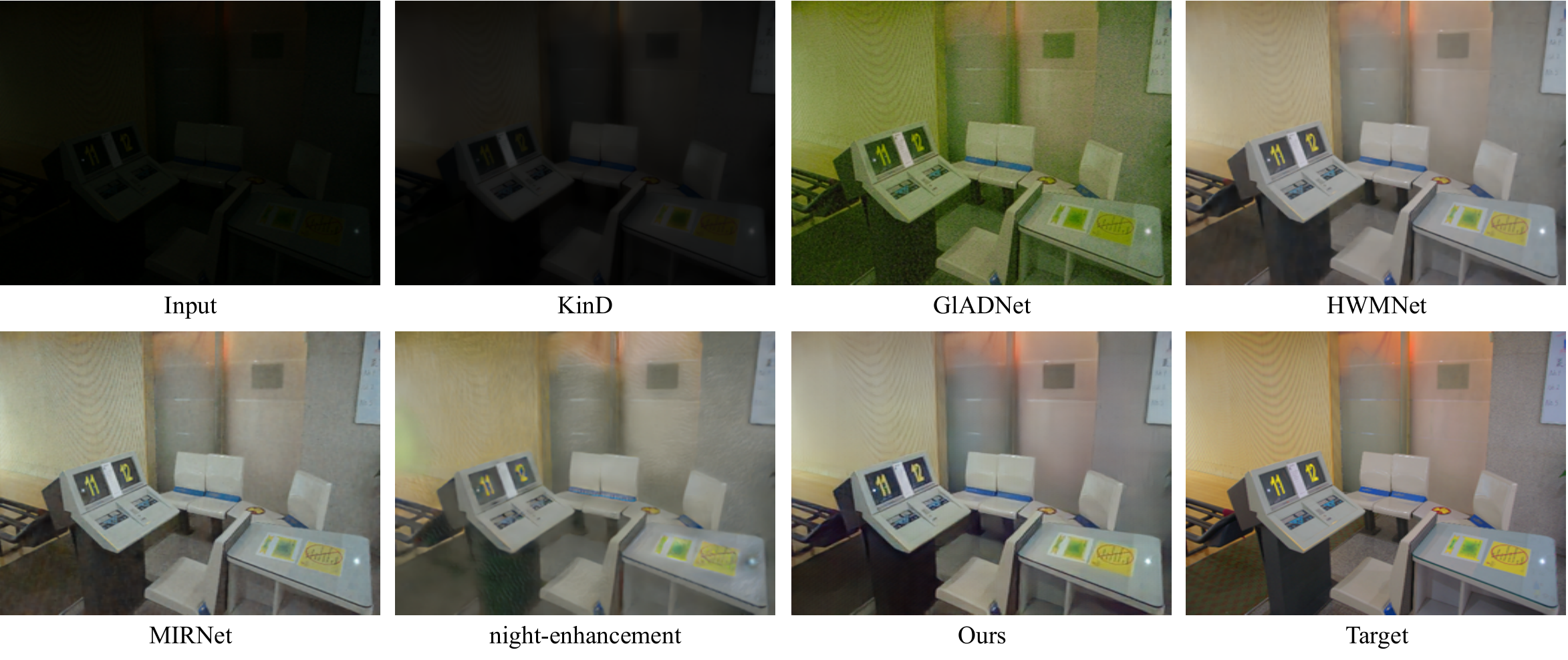}
     \caption{Visual comparison results between Restorer and low-light enhancement baselines on LOL.}
     \label{Fig11}
   \end{figure*}
\begin{figure*}[!t]
    \centering
    \setlength{\abovecaptionskip}{1mm} 
    \setlength{\belowcaptionskip}{-1mm}
    \includegraphics[width=0.93\linewidth]{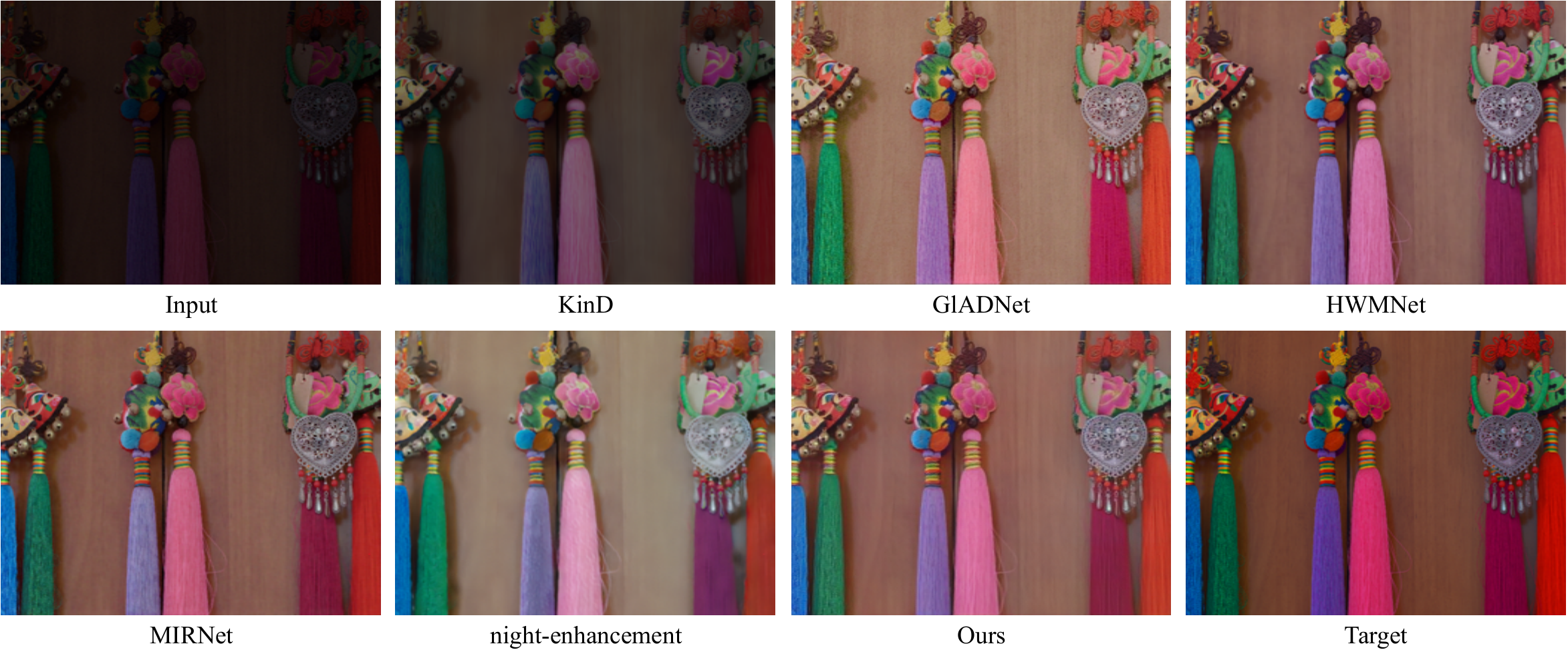}
     \caption{Visual comparison results between Restorer and low-light enhancement baselines on LOL.}
     \label{Fig11(1)}
   \end{figure*}

\paragraph{Low-light Enhancement.} We show qualitative comparison results between Restorer and current state-of-the-art low-light enhancement algorithms in Figure~\ref{Fig11} and Figure~\ref{Fig11(1)}. It can be observed that our method successfully achieves pleasing visual effects which are closer to the ground truth.


\clearpage

\clearpage
\bibliography{aaai25}
\end{document}